\pdfoutput=1 

\documentclass[10pt,twocolumn,letterpaper]{article}

\usepackage[pagenumbers]{cvpr} % To force page numbers, e.g. for an arXiv version
\usepackage{bbding}
\usepackage{pifont}

\usepackage{tabularx}
\usepackage{multirow} 
\definecolor{cvprblue}{rgb}{0.21,0.49,0.74}

\usepackage[pagebackref,breaklinks,colorlinks,allcolors=cvprblue]{hyperref}

\title{Vision-Language Models for Automated 3D PET/CT Report Generation}

\author{
    Wenpei Jiao$^{1}$\thanks{Equal contribution}, Kun Shang$^{2}$\footnotemark[1], Hui Li$^{3}$, Ke Yan$^{4,5}$, Jiajin Zhang$^{4,5,6}$\\ 
    Guangjie Yang$^{7}$, Lijuan Guo$^{8}$, Yan Wan$^{9}$, Xing Yang$^{2}$\thanks{Corresponding authors}, Dakai Jin$^{4}$\footnotemark[2], Zhaoheng Xie$^{1}$\footnotemark[2] \\[0.5em] 
    $^{1}$Institute of Medical Technology and National Biomedical Imaging Center, Peking University\\
    $^{2}$Peking University People’s Hospital,
    $^{3}$Peking University Third Hospital\\
    $^{4}$DAMO Academy, Alibaba Group, 
    $^{5}$Hupan Lab, 
    $^{6}$Shanghai Jiao Tong University\\
    $^{7}$The Affiliated Hospital of Qingdao University\\
    $^{8}$The First Affiliated Hospital of Henan Medical University\\
    $^{9}$Jiujiang City Key Laboratory of Cell Therapy, Jiu Jiang NO.1 People's Hospital\\
    {\tt\small 
yangxing2017@bjmu.edu.cn, dakai.jin@gmail.com, xiezhaoheng@pku.edu.cn} 
}

\begin{document}
\maketitle

\def\thefootnote{\arabic{footnote}} 
\begin{abstract}
Positron emission tomography/computed tomography (PET/CT) is essential in oncology, yet the rapid expansion of scanners has outpaced the availability of trained specialists, making automated PET/CT report generation (PETRG) increasingly important for reducing clinical workload. Compared with structural imaging (e.g., X-ray, CT, and MRI), functional PET poses distinct challenges: metabolic patterns vary with tracer physiology, and whole-body 3D contextual information is required rather than local-region interpretation. To advance PETRG, we propose PETRG-3D, an end-to-end 3D dual-branch framework that separately encodes PET and CT volumes and incorporates style-adaptive prompts to mitigate inter-hospital variability in reporting practices. We construct PETRG-Lym, a multi-center lymphoma dataset collected from four hospitals (824 reports w/ 245,509 paired PET/CT slices), and construct AutoPET-RG-Lym, a publicly accessible PETRG benchmark derived from open imaging data but equipped with new expert-written, clinically validated reports (135 cases). To assess clinical utility, we introduce PETRG-Score, a lymphoma-specific evaluation protocol that jointly measures metabolic and structural findings across curated anatomical regions. Experiments show that PETRG-3D substantially outperforms existing methods on both natural language metrics (e.g., +31.49\% ROUGE-L) and clinical efficacy metrics (e.g., +8.18\% PET-All), highlighting the benefits of volumetric dual-modality modeling and style-aware prompting. Overall, this work establishes a foundation for future PET/CT-specific models emphasizing disease-aware reasoning and clinically reliable evaluation. Codes, models, and AutoPET-RG-Lym will be released.
\end{abstract}    
\section{Introduction}
\label{sec:intro}

\begin{figure}[t]
  \centering
  % \fbox{\rule{0pt}{2in} \rule{1.1\linewidth}{0pt}}
   \includegraphics[width=1.\linewidth]{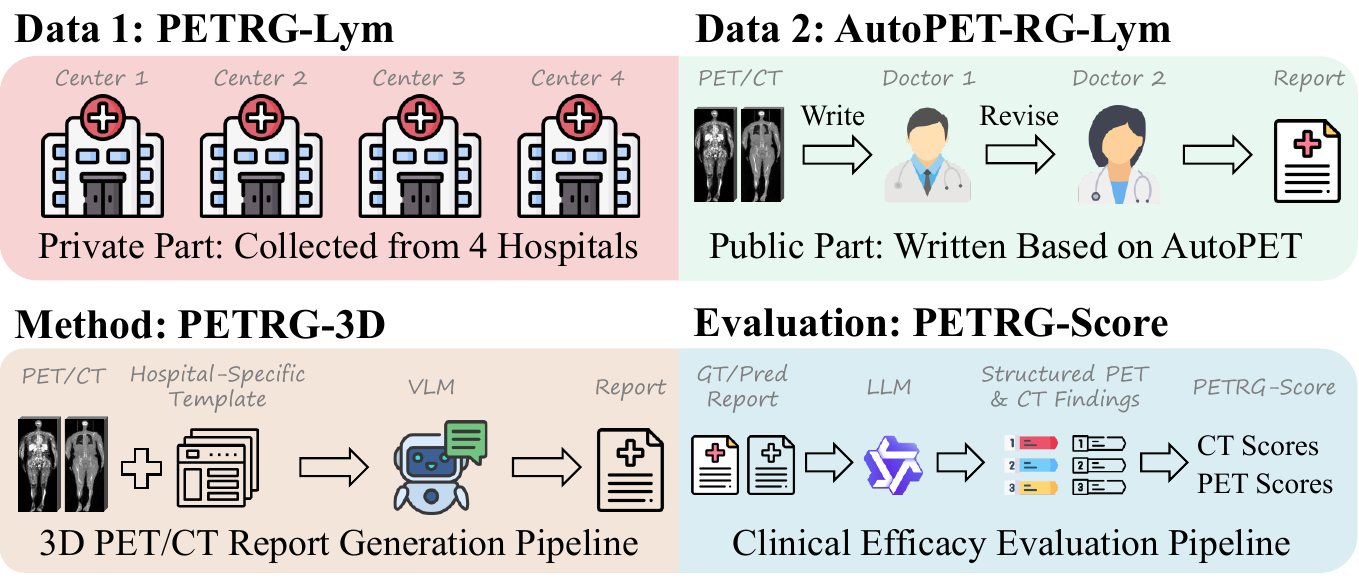}

   \caption{A glimpse of this work. (1) Data: The dataset consists of a private part collected from four medical centers in China and a public part derived from the AutoPET dataset \cite{gatidis2023autopet}, for which reports were generated and subsequently revised and verified by two senior nuclear medicine physicians. (2) Method: An end-to-end 3D PET/CT report-generation framework that integrates dual-modality volumetric encoding and hospital-specific prompting to generate clinically coherent reports. (3) Evaluation: A clinical efficacy evaluation pipeline that quantifies both metabolic (PET) and anatomical (CT) accuracy of generated reports.}
   \label{fig:fig1}
\end{figure}

Intelligent interpretation of medical imaging has achieved notable progress in recent years, especially for structural modalities such as X-ray and CT \cite{hyland2023maira,tanno2025collaboration,tanida2023interactive,gu2024complex,bannur2024maira,jin2025chain,jin2024promptmrg}. In contrast, positron emission tomography/computed tomography (PET/CT) provides high-sensitivity, quantitative functional information such as glucose metabolism. This complementary information is essential for cancer diagnosis, staging, treatment planning, and therapy response assessment \cite{zhang2025pet2rep,gao2025hierarchical,yu2025location}. Consequently, PET/CT has become widely adopted in oncology \cite{huemann2025vision,nguyen2025toward,tie2024personalized}.

PET/CT report writing is a core task in nuclear medicine, yet the rapid increase in installed scanners and scan volumes has intensified clinical workload. Compounded by the persistent shortage of trained nuclear medicine physicians, this creates a growing gap between imaging demand and available diagnostic capacity\cite{gao2025hierarchical,zhang2025pet2rep}. Automated PET/CT report generation (PETRG) therefore holds strong potential to alleviate this pressure and improve reporting efficiency.

\begin{table*}[htbp]
\small
  \centering
  \caption{Comparison of benchmarks for PETRG task. ``Mul-Cnt'', ``Dis-Spe'', ``FT'', ``Info'', ``US Num'', and ``AL Num'' are short for Multi-Center, Disease-Specific, Fine-Tune, Information, the Number of Uptaking Status, and the Number of Anatomical Locations, respectively.}
  \resizebox{2.\columnwidth}{!}{
    \begin{tabular}{lrcccccccccccc}
    \toprule
    \multicolumn{1}{c}{\multirow{2}[3]{*}{\textbf{Benchmarks}}} & \multicolumn{1}{c}{\multirow{2}[3]{*}{\textbf{Venues}}} & \multicolumn{4}{c}{\textbf{Datasets}} & \multicolumn{4}{c}{\textbf{Methods}} & \multicolumn{4}{c}{\textbf{Metrics}} \\
\cmidrule{3-5}\cmidrule{7-9}\cmidrule{11-14}          &       & \textbf{Mul-Cnt} & \textbf{Dis-Spe} & \textbf{Size} &       & \textbf{Input} & \textbf{3D} & \textbf{FT} &       & \textbf{CT Info} & \textbf{PET Info} & \textbf{US Num} & \textbf{AL Num} \\
\midrule
\midrule
    ViMed-PET\cite{nguyen2025toward} & \multicolumn{1}{c}{NeurIPS'25} & \ding{55} & \ding{55} & 2757  &       & PET   & \checkmark & \checkmark &       & \checkmark & \checkmark & 2     & 5 \\
    PET2Rep\cite{zhang2025pet2rep} & \multicolumn{1}{c}{AAAI'26} & \ding{55} & \ding{55} & 565   &       & PET, CT & \ding{55} & \ding{55} &       & \ding{55} & \checkmark & 4     & 19 \\
    PETRG (Ours) & \multicolumn{1}{c}{-}      & \checkmark & \checkmark & 959   &       & PET, CT & \checkmark & \checkmark &       & \checkmark & \checkmark & 5     & 24 \\
    \bottomrule
    \end{tabular}%
}
  \label{tab:tab1}%
\end{table*}%

Metabolic patterns in PET vary substantially with tracer physiology, making functional interpretation more complex than that of structural imaging. Second, PET/CT in oncology typically requires whole-body 3D contextual reasoning rather than localized-region (e.g., chest, brain) analysis, as physicians must jointly describe anatomical structures and metabolic activity across the entire body \cite{zhang2025pet2rep}. These factors drastically increase the difficulty of automated report generation. Finally, obtaining large-scale paired PET/CT image–report datasets is challenging, as PET/CT examinations are more costly, less frequently performed than X-ray or CT, and require expert-validated reports, limiting data availability for model development.

Recent advances in Vision-Language Models (VLMs) have significantly advanced report generation for single-modality 2D/3D medical images (e.g., X-ray, CT) \cite{bannur2024maira, jin2024promptmrg, hamamci2024ct2rep, lai2024e3d}. However, these approaches are
inherently limited when applied to dual-modality 3D PET/CT imaging. Consequently, automated report generation for complete 3D PET/CT scans remains a critical and unaddressed challenge. To address the aforementioned challenges, we investigate from three perspectives: model, datasets, and evaluation metrics. Figure~\ref{fig:fig1} illustrates our contributions in these aspects, and Table~\ref{tab:tab1} summarizes the comparison with prior work.

At the model level, inspired by recent advances in 3D CT report generation (CTRG) with VLMs~\cite{chen2024dia,wu2023towards,chen2025large}, we propose an end-to-end 3D PETRG framework. The architecture comprises three key components: (1) Dual-Stream Volumetric Feature Encoding: a two-branch 3D visual encoder separately captures functional uptake information from PET and anatomical density information from CT volumes. (2) Style-Adaptive Multimodal Fusion: hospital-specific prompt templates are introduced to address inter-center variation in reporting style, dynamically tailoring prompts to institutional conventions. (3) Parameter-Efficient Report Generation: the fused multimodal embeddings are fed into a language decoder to produce clinically coherent reports. In contrast to prior PETRG works—PET2Rep~\cite{zhang2025pet2rep}, which processes only 2D slices with off-the-shelf VLMs, and ViMed-PET~\cite{nguyen2025toward}, who utilize only 3D PET without CT—our model is the first to jointly model full 3D PET and CT volumes in an end-to-end manner.

At the data level, we collect and curate 824 PET/CT scan–report pairs from 746 lymphoma patients collected across four medical centers in China, forming PETRG-Lym, the first large-scale, multicenter, single-disease PETRG dataset. Additionally, we identify 135 lymphoma cases from the public AutoPET dataset \cite{gatidis2023autopet}, for which two senior nuclear medicine physicians generated and refined structured reports, yielding AutoPET-RG-Lym, an open external benchmark for PETRG evaluation and cross-center validation.

At the evaluation level, beyond standard natural language generation (NLG) metrics such as BLEU~\cite{papineni2002bleu}, METEOR~\cite{banerjee2005meteor}, and ROUGE~\cite{lin2004rouge}, we introduce PETRG-Score, a clinically driven metric developed with senior nuclear medicine physicians. PETRG-Score explicitly evaluates the model’s ability to accurately describe both PET uptake patterns and CT structural findings, ensuring clinical fidelity beyond text similarity.

Using the proposed benchmark, we train and validate our method on the PETRG-Lym dataset and test it externally on AutoPET-RG-Lym. Extensive experiments reveal that current state-of-the-art VLMs for CTRG perform poorly when directly applied to multi-center PET/CT data, revealing a substantial domain gap. In contrast, our proposed PETRG-3D framework—featuring dual-branch volumetric encoding and hospital-specific prompting—achieves significant improvements in both language quality and clinical fidelity. These results highlight the necessity of PETRG datasets and multimodal architectures for functional imaging report generation.

Our main contributions are summarized as follows: 
\begin{itemize}
    \item We propose a 3D PETRG model that leverages a dual-branch architecture to separately encode volumetric PET and CT features, along with a hospital-style-aware prompting strategy for multi-center report style adaptation.
    \item We construct PETRG-Lym, a multi-center lymphoma PETRG dataset collected from four hospitals (824 scans, 245,509 paired slices), and we release AutoPET-RG-Lym, a publicly accessible benchmark derived from open PET/CT images but furnished with new expert-written, clinically validated reports (135 cases). 
    \item PETRG-Score, a new evaluation metric that jointly assesses metabolic uptake and structural information in PET/CT reports, incorporating fine-grained anatomical grounding.
    \item Extensive experiments demonstrate the effectiveness of our method, reveal limitations of existing paradigms, and establish a strong foundation for future research in 3D PETRG.
\end{itemize}

\section{Related Works}
\label{sec:related works}

Our work is most relevant to the growing literature on X-Ray, CT, MRI and PETRG.

\subsection{X-Ray Report Generation}
Recent advances in 2D medical report generation have been driven by the availability of large-scale public datasets and progress in multimodal learning. Benchmark datasets such as IU-Xray~\cite{demner2015preparing} and MIMIC-CXR~\cite{johnson2019mimic} have provided essential data foundations, while the evolution of deep learning—particularly VLMs—has spurred the development of numerous sophisticated report generation frameworks~\cite{jin2024promptmrg,jin2025chain,chen2020generating,nooralahzadeh2021progressive,yang2022knowledge,yang2023radiology,wang2023metransformer,huang2023kiut,li2023dynamic}.

Most approaches follow an encoder–decoder architecture, where visual features extracted by CNNs or vision transformers are decoded into diagnostic narratives (e.g., MAIRA-1~\cite{hyland2023maira} and Flamingo-CXR~\cite{tanno2025collaboration}). However, these 2D X-ray–based methods are not directly applicable to PET/CT, as they disregard 3D volumetric context essential for accurate interpretation of metabolic activity and structural relationships.

\subsection{CT \& MRI Report Generation}

Research in 3D medical report generation (CT/MRI)~\cite{hamamci2024foundation} is dominated by 3D vision encoder-language decoder architectures~\cite{bai2024m3d, lai2024e3d, li2025towards}, often built upon pre-trained encoders like CT-CLIP~\cite{hamamci2024foundation}, RadFM~\cite{wu2023towards}, or M3D~\cite{bai2024m3d}. To manage 3D data complexity, methods employ diverse strategies. Some integrate memory modules, such as the disease prototype memory in Dia-LLaMA~\cite{chen2024dia} for class imbalance or the relational memory in CT2Rep~\cite{hamamci2024ct2rep} for coherence. Others leverage fine-grained anatomical priors\cite{lin2024ct}, as seen in Reg2RG~\cite{chen2025large} which utilizes explicit correspondences from RadGenome-ChestCT~\cite{zhang2024radgenome}. For MRI, AutoRG-Brain~\cite{lei2024autorg} fuses global and local context using automatic ROIs.

However, existing 3D CT/MRI benchmarks are single-modality, lacking the cross-modality reasoning required for PET/CT, which demands joint interpretation of metabolic and anatomical information.

\subsection{PET/CT Report Generation}
Efforts toward automated PETRG have emerged only recently\cite{tie2024personalized,gao2025hierarchical,yu2025location,choi2025empowering}. PET2Rep~\cite{zhang2025pet2rep} first benchmarked the task by applying VLMs to 2D PET/CT slices with prompt engineering revealing that both general-purpose and medical-specific VLMs perform poorly on PETRG. ViMed-PET~\cite{nguyen2025toward} introduced a Vietnamese PET/CT report and VQA dataset, proposing a three-stage fine-tuning pipeline for language adaptation. However, both approaches suffer from significant limitations. PET2Rep~\cite{zhang2025pet2rep} process only a handful of 2D slices, discarding rich 3D spatial context inherent in PET/CT volumes. ViMed-PET~\cite{nguyen2025toward} use PET images alone, neglecting CT structural cues. These limitations underscore the need for approaches that jointly model full 3D PET and CT volumes, leveraging complementary metabolic and anatomical information.

\section{Multi-Center, Single-Disease PET/CT Report Generation Dataset}

\begin{table*}[htbp]
\small
  \centering
  \caption{Statistics of PETRG-Lym and AutoPET-RG-Lym. S Num and P Num denote the number of scans and patients, respectively.}
    \resizebox{2.05\columnwidth}{!}{
    \begin{tabular}{l|c|ccc|ccc|cc}
    \toprule
    \multirow{2}[4]{*}{\textbf{Dataset}} & \multirow{2}[4]{*}{\textbf{Center ID}} & \multicolumn{3}{c|}{\textbf{Training Cohort}} & \multicolumn{3}{c|}{\textbf{Validation Cohort}} & \multicolumn{2}{c}{\textbf{Total}} \\
\cmidrule{3-10}          &       & \textbf{P Num} & \textbf{S Num} & \textbf{Scan Period} & \textbf{P Num} & \textbf{S Num} & \textbf{Scan Period} & \textbf{P Num} & \textbf{S Num} \\
    \midrule
    \midrule
    \multirow{4}[1]{*}{PETRG-Lym} & 1     & 224   & 227   & 2023.10 - 2024.07 & 57    & 57    & 2024.07 - 2024.09 & 281   & 284 \\
          & 2     & 107   & 147   & 2015.07 - 2024.03 & 28    & 36    & 2024.05 - 2025.05 & 135   & 183 \\
          & 3     & 112   & 138   & 2018.01 - 2024.03 & 29    & 30    & 2024.03 - 2025.01 & 141   & 168 \\
          & 4     & 151   & 151   & 2013.02 - 2019.05 & 38    & 38    & 2019.05 - 2021.04 & 189   & 189 \\
    \midrule
    AutoPET-RG-Lym & 5   & -     & -     & -     & 135   & 135   & -     & 135   & 135 \\
    \midrule
    Total & -     & 594   & 663   & - & 287   & 296   & - &  881   & 959 \\
    \bottomrule
    \end{tabular}%
  \label{tab:tab 2}%
  }
\end{table*}%

\subsection{Motivation}  

Progress in PETRG is hindered by the limited availability of paired PET/CT image–report datasets. Existing PET datasets (e.g., AutoPET~\cite{gatidis2023autopet}, HECKTOR ~\cite{andrearczyk2022head}) commonly lack reports, and recent PETRG resources are single-center and multi-disease (Table~\ref{tab:tab1}), introducing substantial heterogeneity in imaging appearance and clinical wording that complicates modeling and increases the risk of diagnostic errors. In addition, the field lacks a standardized benchmark for fair evaluation, as no public PET/CT dataset currently provides clinically validated reports.

To address these limitations, we focus on a single disease while collecting data from multiple centers. This design reduces cross-disease heterogeneity while still capturing real-world variability, supporting a practical “clinician confirms, AI drafts” workflow for reliable PETRG. Furthermore, we construct a publicly accessible PET/CT report-generation benchmark equipped with new expert-written, clinically validated reports, enabling reproducible and consistent evaluation for future work.

\subsection{The PETRG-Lym Dataset}  

The PETRG-Lym dataset is sourced from the nuclear medicine departments of four hospitals in China, spanning cities of varying scale to ensure diversity and representativeness. All PET/CT scans were acquired using the \textsuperscript{18}F-FDG radiotracer without contrast-enhanced CT. Images are preserved in their original DICOM format and include essential metadata such as patient height, weight, age, radiotracer injection time, and dose. Imaging reports were authored by nuclear medicine physicians and subsequently reviewed by senior staff, covering patient demographics, clinical history, acquisition protocol, findings, and diagnostic impressions. In this work, we focus exclusively on the ``findings'' section. As summarized in Table \ref{tab:tab 2}, the PETRG-Lym dataset comprises 824 PET/CT–report pairs from 746 lymphoma patients across the four centers, collected over a 12-year period from February 2013 to May 2025. This constitutes the first large-scale, multi-center, single-disease PETRG dataset.

\subsection{The AutoPET-RG-Lym Dataset}  
To enable external validation and promote community benchmarking, we construct AutoPET-RG-Lym from lymphoma cases in the public AutoPET dataset~\cite{gatidis2023autopet}, which was collected at University Hospital Tübingen and University Hospital of the LMU Munich. It contains patients with lymphoma, malignant melanoma, and lung cancer. This dataset differs from our internal data in three key aspects: (1) it originates from a different country, (2) it was acquired using different imaging hardware, and (3) it employs different CT contrast agents—making it a strong candidate for external validation of model generalization.

% Given the current lack of publicly available PETRG benchmarks for fair community evaluation, w

We carefully selected 135 lymphoma cases from AutoPET~\cite{gatidis2023autopet} and commissioned two senior nuclear medicine physicians from top-tier Chinese hospitals to independently compose structured reports. These reports underwent cross-review and iterative refinement to ensure high clinical fidelity. The resulting dataset provides an open validation benchmark for PETRG research.

\subsection{Data Preprocessing}  
Starting from de-identified DICOM data, we extracted patient metadata (e.g., weight, radiotracer dose, injection time) and converted PET and CT volumes to NIfTI format. All volumes were reoriented to RAS and resampled to a uniform spacing of $1.5 \times 1.5 \times 3$ mm, with CT matched to PET dimensions. CT voxel intensities were converted to Hounsfield Units (HU) and clipped to $[-1000, +1000]$. Standardized uptake value (SUV) normalization was performed using the extracted metadata. We employed TotalSegmentator~\cite{wasserthal2023totalsegmentator} to remove the scanning bed from CT backgrounds. Given the presence of both “head-to-midthigh” and “head-to-toe” scan protocols—and the observation that reports seldom describe regions below the mid-thigh—we uniformly cropped all volumes at the upper thigh. Additional preprocessing details, including voxel intensity standardization, scanning bed removing, body part cropping, and report text cleaning, are provided in the supplementary material.

\subsection{Dataset Construction and Splitting}  
We split PETRG-Lym into training and internal validation sets chronologically, as shown in Table \ref{tab:tab 2}. To prevent data leakage, we grouped scans by patient and, within each center, assigned the first 80\% of patients (by initial scan date) to training and the remaining 20\% to validation. This yields 663 scans (594 patients) for training and 161 scans (152 patients) for validation. All 135 samples in AutoPET-RG-Lym serve as an external test set.

\section{Methods}
% --- FIGURE (NO CHANGE NEEDED) ---
\begin{figure*}[t]
  \centering
  \includegraphics[width=0.97\linewidth]{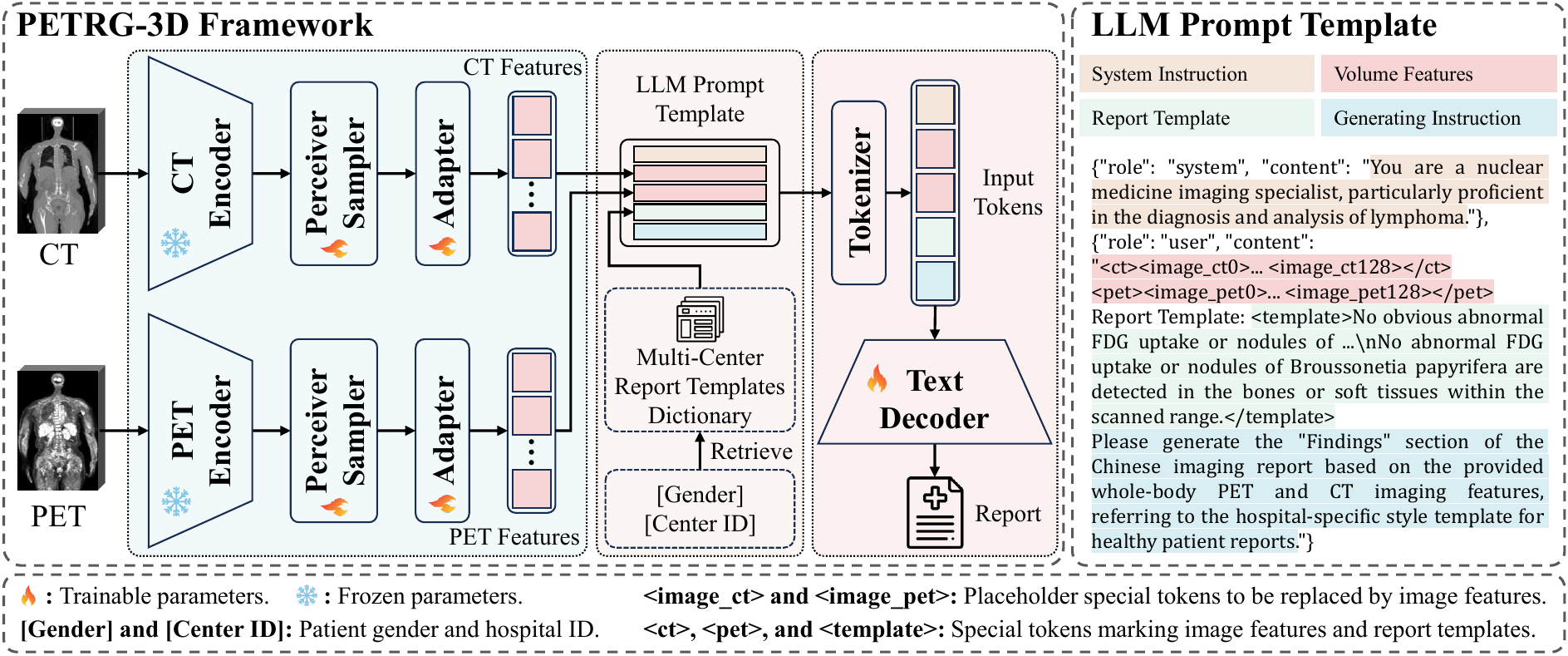}
  \caption{Overall framework of PETRG-3D. The model comprises three key components: (1) a dual-stream 3D visual feature extractor (DSFE) that processes PET and CT volumes separately \textit{(blue)}; (2) a style-adaptive multimodal fusion (SAMF) module that dynamically integrates visual features with hospital-specific prompt templates \textit{(yellow)}; and (3) a LoRA-adapted large language model (LLM) for effective radiology report generation \textit{(pink)}.}
  \label{fig:fig3}
\end{figure*}
% --- END FIGURE ---

Inspired by recent CT report generation frameworks~\cite{wu2023towards,bai2024m3d}, we propose PETRG-3D, a novel pipeline for 3D PET/CT report generation. As illustrated in Figure~\ref{fig:fig3}, our approach consists of three key stages: 
(1) \textbf{Dual-Stream Volumetric Feature Encoding (DSFE)} from PET and CT scans; 
(2) \textbf{Style-Adaptive Multimodal Fusion (SAMF)} using hospital-specific prompt templates; and 
(3) parameter-efficient autoregressive report generation using a Low-Rank Adaptation (LoRA)-adapted LLM. 
We detail each component below.

\subsection{Dual-Stream Volumetric Feature Encoding}

To jointly model metabolic (PET) and anatomical (CT) information, we propose a dual-stream 3D visual encoder. Unlike prior approaches using 2D slices~\cite{zhang2025pet2rep} or single-modality 3D PET~\cite{nguyen2025toward}, our \textbf{DSFE} architecture (Fig.~\ref{fig:fig3}, blue region) explicitly processes and fuses these complementary volumetric modalities.

Each stream comprises a volume encoder followed by a perceiver sampler. The volume encoder employs a sliding-window strategy to process input volumes of arbitrary size, producing high-dimensional feature maps. The perceiver sampler then compresses these variable-length features into a fixed-length sequence of representative embeddings.

To leverage rich, pre-learned representations, we initialize both volume encoders with the 3D ViT weights pretrained on large-scale CT data (RadFM~\cite{wu2023towards}) and keep them frozen. This strategy is highly parameter-efficient. However, a significant domain gap exists between RadFM's lung-centric CT data and our whole-body PET/CT scans. To bridge this gap, we retrain \textit{only} the perceiver samplers on our PETRG-Lym dataset, allowing them to adapt the frozen features to our specific PET and CT modalities independently.

\subsection{Style-Adaptive Multimodal Fusion}

The compressed PET and CT feature sequences from the DSFE module are projected into the LLM's hidden space via linear transformations, aligning the visual and linguistic semantic spaces.

A key challenge in multicenter report generation is the significant stylistic heterogeneity across institutions. Our \textbf{SAMF} module addresses this by emulating the clinical workflow. We curate a dictionary of hospital-specific ``healthy-example'' templates. These templates also account for gender (male/female) to handle anatomical differences, particularly in pelvic descriptions.

As shown in Figure~\ref{fig:fig3} (yellow region), each sample carries two metadata flags: [Center ID] and [Gender]. These flags dynamically retrieve the corresponding hospital- and gender-matched template from the dictionary, which is then injected into the input prompt. To guide the model in distinguishing these inputs, we wrap the CT features, PET features, and template text with dedicated special tokens (\textless ct\textgreater, \textless pet\textgreater, and \textless template\textgreater, respectively).

\subsection{Parameter-Efficient Autoregressive Generation}

The final fused multimodal embedding sequence—comprising visual tokens and the template-aware textual prompt—is fed directly into a pre-trained LLM (e.g., Qwen3-8B~\cite{yang2025qwen3}) for autoregressive report generation (Fig.~\ref{fig:fig3}, pink region).

Instead of full fine-tuning, we adopt LoRA~\cite{hu2022lora} for parameter-efficient training. LoRA injects trainable, low-rank matrices into the Transformer's attention layers while freezing the original LLM parameters. This strategy enables the model to learn PET/CT-specific terminology and reporting structures with minimal trainable parameters, mitigating catastrophic forgetting and substantially reducing computational overhead.

\subsection{Implementation and Evaluation Protocol}
\subsubsection{Evaluation Metrics}
We evaluate performance on two axes: NLG and Clinical Efficacy (CE). 

\begin{table*}[htbp]
\small
  \centering
  \caption{Evaluation of NLG and CE metrics on the PETRG-Lym dataset. PET2Rep-Sep and PET2Rep-Fus refer to the two image input strategies from~\cite{zhang2025pet2rep}: ``Sep'' denotes separate PET and CT slice inputs, while ``Fus'' denotes a fused PET/CT image input. ``All'' and ``Ab'' denote results on all and only abnormal classes.}
  \resizebox{2.1\columnwidth}{!}{
    \begin{tabular}{l|l|cccccc|cccc}
    \toprule
    {\multirow{2}[4]{*}{\textbf{Methods}}} & \multirow{2}[4]{*}{\textbf{LLMs/VLMs}} & \multicolumn{6}{c|}{\textbf{NLG Metrics}}     & \multicolumn{4}{c}{\textbf{CE Metrics}} \\
\cmidrule{3-12}          &       & \textbf{B-1} & \textbf{B-2} & \textbf{B-3} & \textbf{B-4} & \textbf{MTR} & \textbf{R-L} & \textbf{PT-All} & \textbf{CT-All} & \textbf{PT-Ab} & \textbf{CT-Ab} \\
    \midrule
    \midrule
    LLaVA-Med\cite{nguyen2024multilingual} & Mistral-7B-v0.2 & 3.21  & 0.99  & 0.56  & 0.40  & 2.46  & 4.84  & -     & -     & -     & - \\
    RadFM\cite{wu2023towards} & Llama2-13B & 1.65  & 0.38  & 0.21  & 0.16  & 1.27  & 2.44  & -     & -     & -     & - \\
    M3D\cite{bai2024m3d}   & Llama2-7B & 0.53  & 0.21  & 0.11  & 0.07  & 2.16  & 4.56  & -     & -     & -     & - \\
    \multirow{3}[0]{*}{PET2Rep-Sep\cite{zhang2025pet2rep}} & Qwen3-VL-8B & 39.60  & 21.61  & 11.39  & 6.16  & 27.06  & 22.25  & 23.88  & 33.66  & 11.53  & 28.41  \\
          & InternVL-3.5-8B & 38.23  & 20.74  & 10.96  & 5.96  & 25.99  & 22.20  & 22.75  & 33.45  & 10.64  & 28.01  \\
          & Qwen2.5-VL-Max & 34.98  & 18.85  & 9.97  & 5.48  & 27.91  & 20.37  & 21.78  & 23.93  & 7.84  & 17.11  \\
    \multirow{3}[1]{*}{PET2Rep-Fus\cite{zhang2025pet2rep}} & Qwen3-VL-8B & 38.98  & 20.94  & 10.99  & 5.96  & 27.12  & 21.77  & 23.82  & 30.61  & 11.34  & 24.65  \\
          & InternVL-3.5-8B & 37.79  & 20.44  & 10.78  & 5.83  & 25.69  & 21.92  & 23.88  & 33.88  & 12.22  & \underline{28.62}  \\
          & Qwen2.5-VL-Max & 34.30  & 18.57  & 9.87  & 5.42  & 28.55  & 19.84  & 22.88  & 26.95  & 9.54  & 20.60  \\
    \midrule
    \multirow{6}[2]{*}{PETRG-3D (Ours)} & Llama2-7B & 28.19  & 21.84  & 17.72  & 14.90  & 22.60  & 27.19  & 26.10  & 25.45  & 11.99  & 17.97  \\
          & Mistral-7B-v0.3 & 53.18  & 45.53  & 40.12  & 36.07  & 43.20  & 48.60  & 27.25  & 30.24  & 13.52  & 23.55  \\
          & Qwen2.5-7B & \underline{60.42}  & \underline{52.00}  & \underline{45.92}  & 41.33  & \textbf{ 51.21 } & 52.15  & \underline{31.78}  & \underline{34.13}  & \underline{19.20}  & 28.06  \\
          & Gemma2-9B & 58.57  & 50.24  & 44.25  & 39.79  & 49.47  & 51.49  & 28.29  & 31.92  & 15.08  & 25.54  \\
          & GLM4-9B & 58.73  & 51.22  & 45.72  & \underline{41.55}  & 50.80  & \textbf{ 53.74 } & 29.06  & 31.65  & 15.72  & 25.07  \\
          & Qwen3-8B & \textbf{ 60.78 } & \textbf{ 52.45 } & \textbf{ 46.42 } & \textbf{ 41.90 } & \underline{51.16}  & \underline{52.88}  & \textbf{ 32.06 } & \textbf{ 34.76 } & \textbf{ 19.53 } & \textbf{ 28.76 } \\
    \bottomrule
    \end{tabular}%
}
  \label{tab:tab 3}%
\end{table*}%

\textbf{For NLG metrics}, we report standard metrics: BLEU (B-n)~\cite{papineni2002bleu}, METEOR (MTR)~\cite{banerjee2005meteor}, and ROUGE-L (R-L)~\cite{lin2004rouge}, to quantify textual similarity between generated and reference reports.

\textbf{For CE metrics}, We implement an LLM-based pipeline to extract diagnostic content from reports across 24 lymphoma-specific anatomical regions predefined by senior radiologists. Specifically, we extract labels from reference and generated reports using Qwen3-Max~\cite{yang2025qwen3}, while reference report labels underwent expert validation by a researcher specializing in medical imaging. Labels encompass 5 PET uptake categories (e.g., markedly abnormal, physiological) and 8 CT structural findings (e.g., lymphadenopathy). We evaluate Macro-F1 scores for: \textbf{PET-All}/\textbf{CT-All} (all labels; overall quality) and \textbf{PET-Abnormal (PT-Ab)}/\textbf{CT-Abnormal (CT-Ab)} (abnormal labels only; diagnostic relevance). Full implementation details are in the supplementary material.

\subsubsection{Baselines and Implementation}
We benchmark against strong baselines: LLaVA-Med~\cite{nguyen2024multilingual}, RadFM~\cite{wu2023towards}, and M3D~\cite{bai2024m3d} (prior 2D/3D radiology VLMs), and PET2Rep~\cite{zhang2025pet2rep}, a recent 2D PET/CT method. 
Our model's (PETRG-3D) volume encoder is initialized from the pretrained ViT-3D of RadFM~\cite{wu2023towards}. All experiments are conducted on NVIDIA A800 (80GB) GPUs. Implementation details for baselines and PETRG-3D are provided in the supplementary material.

\section{Experimental Results}  
We evaluate our PETRG-3D framework on the PETRG-Lym and AutoPET-RG-Lym datasets, comparing against existing methods and analyzing each proposed component. Additional results appear in the supplementary material.

\subsection{Main Results}
\subsubsection{Quantitative Results}

Table~\ref{tab:tab 3} details results on PETRG-Lym. Baselines like M3D\cite{bai2024m3d}, pretrained on CT volumes, show limited performance due to domain gaps. The prompt-based PET2Rep yields moderate scores (e.g., 6.16 B-4, 23.88 PT-All with Qwen3-VL-8B~\cite{bai2025qwen2}) but fails to capture PET-specific language. In contrast, our fine-tuned PETRG-3D substantially boosts both NLG (41.9 B-4, 51.16 MTR) and CE metrics (32.06 PT-All, 34.76 CT-All, with Qwen3-8B~\cite{yang2025qwen3})), demonstrating consistent improvements and superior alignment with the radiologists' reporting style.

On the external AutoPET-RG-Lym dataset (Table~\ref{tab:tab 4}), PETRG-3D maintains its superiority in NLG and PET-centric CE metrics, reinforcing the value of domain-specific fine-tuning. However, its CT-centric CE performance degrades. We hypothesize this is due to higher inter-center variance in CT descriptions compared to PET, causing the model to overfit to the source domain's CT style. This suggests limitations of the end-to-end VLMs fine-tuning paradigm. The results underscore that a significant gap remains for the clinical applicability of all current methods.

\begin{table*}[htbp]
\small
  \centering
  \caption{Evaluation of NLG and CE metrics on the External Cohort AutoPET-RG-Lym. PET2Rep-Sep and PET2Rep-Fus refer to the two image input strategies from~\cite{zhang2025pet2rep}.}
  \resizebox{2.1\columnwidth}{!}{
    \begin{tabular}{l|l|cccccc|cccc}
    \toprule
    {\multirow{2}[4]{*}{\textbf{Methods}}} & \multirow{2}[4]{*}{\textbf{LLMs/VLMs}} & \multicolumn{6}{c|}{\textbf{NLG Metrics}}     & \multicolumn{4}{c}{\textbf{CE Metrics}} \\
\cmidrule{3-12}          &       & \textbf{B-1} & \textbf{B-2} & \textbf{B-3} & \textbf{B-4} & \textbf{MTR} & \textbf{R-L} & \textbf{PT-All} & \textbf{CT-All} & \textbf{PT-Ab} & \textbf{CT-Ab} \\
    \midrule
    \midrule
    LLaVA-Med\cite{nguyen2024multilingual} & Mistral-7B-v0.2 & 3.16  & 0.98  & 0.57  & 0.42  & 2.37  & 4.69  & -     & -     & -     & - \\
    RadFM\cite{wu2023towards} & Llama2-13B & 1.57  & 0.45  & 0.24  & 0.18  & 1.12  & 2.14  & -     & -     & -     & - \\
    M3D\cite{bai2024m3d}   & Llama2-7B & 0.28  & 0.09  & 0.04  & 0.03  & 1.84  & 4.33  & -     & -     & -     & - \\
    \multirow{3}[0]{*}{PET2Rep-Sep\cite{zhang2025pet2rep}} & Qwen3-VL-8B & 40.80  & 24.08  & 13.51  & 7.51  & 27.56  & 24.33  & 24.25  & 29.22  & 12.29  & 23.07  \\
          & InternVL-3.5-8B & 37.94  & 22.20  & 12.50  & 6.98  & 26.62  & 23.16  & 23.34  & \textbf{ 33.56 } & 11.78  & \textbf{ 27.60 } \\
          & Qwen2.5-VL-Max & 32.63  & 19.16  & 10.62  & 5.83  & 28.78  & 21.21  & 26.76  & 25.94  & 14.00  & 18.46  \\
    \multirow{3}[1]{*}{PET2Rep-Fus\cite{zhang2025pet2rep}} & Qwen3-VL-8B & 39.36  & 23.19  & 13.00  & 7.20  & 28.00  & 23.40  & 22.62  & 31.17  & 10.61  & 24.96  \\
          & InternVL-3.5-8B & 38.39  & 22.43  & 12.58  & 7.05  & 26.91  & 22.92  & 23.38  & \underline{32.96}  & 12.16  & \underline{27.16}  \\
          & Qwen2.5-VL-Max & 32.13  & 18.97  & 10.56  & 5.82  & 29.12  & 21.01  & 28.31  & 28.57  & 16.14  & 21.46  \\
    \midrule
    \multirow{6}[2]{*}{PETRG-3D (Ours)} & Llama2-7B & 24.05  & 19.82  & 16.87  & 14.68  & 26.31  & 32.20  & 19.02  & 14.73  & 4.30  & 6.04  \\
          & Mistral-7B-v0.3 & \underline{52.37}  & \underline{43.92}  & \underline{38.65}  & \underline{35.03}  & 40.31  & \underline{44.20}  & 23.63  & 16.94  & 11.64  & 9.58  \\
          & Qwen2.5-7B & 47.90  & 37.88  & 31.51  & 27.35  & 36.50  & 37.30  & 30.42  & 21.90  & 19.01  & 15.77  \\
          & Gemma2-9B & 47.27  & 39.23  & 34.28  & 30.97  & 37.09  & 40.21  & 30.03  & 18.56  & 18.49  & 11.15  \\
          & GLM4-9B & \textbf{ 55.89 } & \textbf{ 50.38 } & \textbf{ 46.76 } & \textbf{ 44.18 } & \textbf{ 47.41 } & \textbf{ 55.47 } & \textbf{ 36.22 } & 17.14  & \textbf{ 25.46 } & 9.24  \\
          & Qwen3-8B & 49.41  & 41.30  & 36.07  & 32.47  & \underline{41.50}  & 40.97  & \underline{33.48}  & 22.79  & \underline{22.81}  & 16.35  \\
    \bottomrule
    \end{tabular}%
}
  \label{tab:tab 4}%
\end{table*}%

\subsubsection{Qualitative Results}

\begin{figure*}[t]
  \centering
  \includegraphics[width=0.98\linewidth]{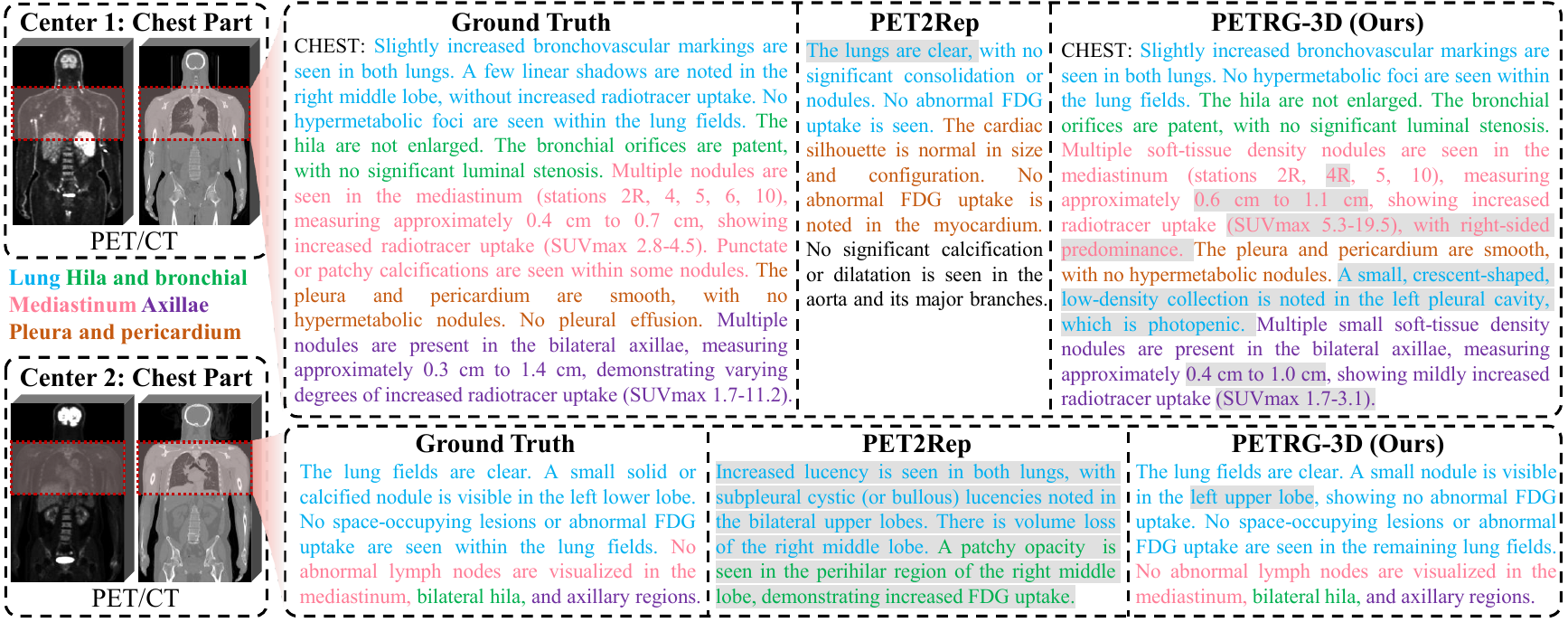}
  \caption{Qualitative comparison of our method against PET2Rep\cite{zhang2025pet2rep} on chest results. Different colors denote distinct anatomical areas. Incorrect diagnoses are highlighted in gray.}
  \label{fig:fig4}
\end{figure*}

Fig.~\ref{fig:fig4} presents a qualitative comparison on PETRG-Lym, showcasing two cases from different centers to highlight diverse reporting styles. We display key findings of the chest region for brevity; distinct colors denote different anatomical areas, and gray backgrounds indicate incorrect diagnoses.

In terms of reporting style, PET2Rep generates an excessively brief and anatomically incomplete report for Case 1. In contrast, our model generates a report whose length and anatomical coverage closely align with the ground truth (GT). In Case 2, while PET2Rep achieves a comparable length, it still exhibits incomplete anatomical coverage. Our model again demonstrates superior adaptability, matching the GT in both length and completeness.

Regarding diagnostic accuracy, PET2Rep incorrectly reports no abnormalities in Case 1. Our model, however, correctly identifies multiple findings, such as ``slightly increased bronchovascular markings''. In Case 2, PET2Rep's output is clinically irrelevant. Our model successfully identifies ``a small nodule... in the left lobe,'' despite a minor error in fine-grained localization (``left upper lobe'' vs. ``left lower lobe'').

\subsection{Ablation Studies and Analysis}
\begin{table*}[htbp]
\small
  \centering
  \caption{Ablation study on the PETRG-Lym dataset. CT Base and PET Base use only CT or PET stream, respectively. DSFE denotes Dual-Stream Volumetric Feature Encoding; SAMF denotes Style-Adaptive Multimodal Fusion.}
   \resizebox{2.05\columnwidth}{!}{ 
    \begin{tabular}{cccc|cccccc|cccc}
    \toprule
    \multirow{2}[4]{*}{\textbf{CT Base}} & \multirow{2}[4]{*}{\textbf{PET Base}} & \multirow{2}[4]{*}{\textbf{DSFE}} & \multirow{2}[4]{*}{\textbf{SAMF}} & \multicolumn{6}{c|}{\textbf{NLG Metrics}}     & \multicolumn{4}{c}{\textbf{CE Metrics}} \\
\cmidrule{5-14}          &       &       &       & \textbf{B-1} & \textbf{B-2} & \textbf{B-3} & \textbf{B-4} & \textbf{MTR} & \textbf{R-L} & \textbf{PT-All} & \textbf{CT-All} & \textbf{PT-Ab} & \textbf{CT-Ab} \\
    \midrule
    \midrule
    \checkmark     &       &       &       & 49.19  & 39.91  & 33.58  & 29.13  & 40.91  & 41.62  & 28.81  & 32.96  & 15.77  & 26.86  \\
          & \checkmark     &       &       & 52.74  & 43.01  & 36.37  & 31.76  & 42.65  & 43.61  & 26.01  & 32.95  & 12.40  & 26.91  \\
          & \checkmark     & \checkmark     &       & 58.02  & 48.30  & 41.47  & 36.54  & 46.88  & 46.93  & 28.38  & 33.27  & 15.36  & 27.41  \\
          & \checkmark     & \checkmark     & \checkmark     & \textbf{60.78} & \textbf{52.45} & \textbf{46.42} & \textbf{41.90} & \textbf{51.16} & \textbf{52.88} & \textbf{32.06} & \textbf{34.76} & \textbf{19.53} & \textbf{28.76} \\
    \bottomrule
    \end{tabular}%
  \label{tab:tab 5}%
  }
\end{table*}%

We analyze the impact of our core components and design choices, using Qwen3-8B~\cite{yang2025qwen3} as the text decoder.

\subsubsection{Impact of Key Components}
We first ablate our two key modules, DSFE and SAMF. Results are in Table~\ref{tab:tab 5}.

\textbf{DSFE:} Jointly encoding CT and PET volumes with DSFE consistently outperforms single-modality inputs (PET-only or CT-only) across nearly all NLG and CE metrics.
\textbf{SAMF:} Adding SAMF further boosts NLG performance substantially. By explicitly incorporating hospital-specific templates, SAMF simplifies the stylistic modeling task. This frees model capacity to focus on clinically relevant details, thereby improving CE metrics as well.

\subsubsection{Analysis of Design Choices}
\begin{figure}[t]
  \centering
  \includegraphics[width=1.\linewidth]{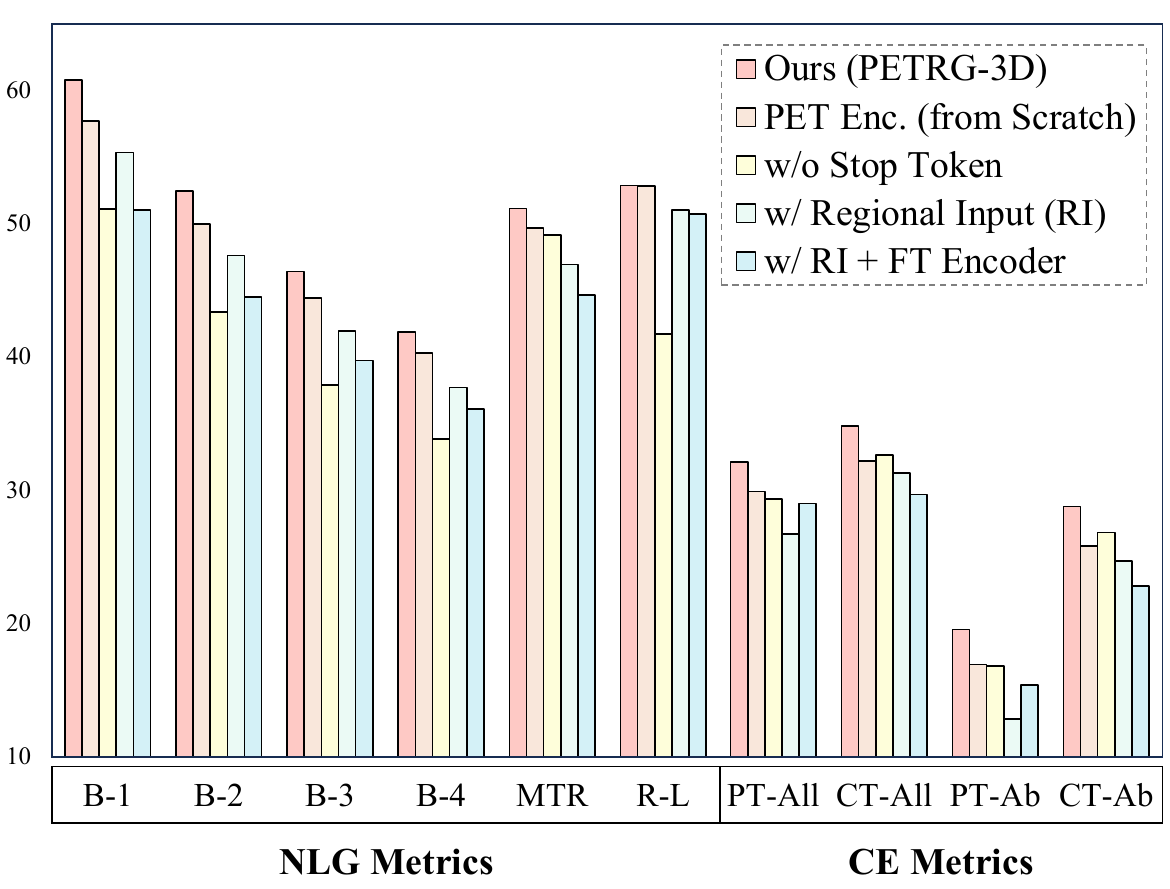}
  \caption{
    \textbf{Analysis of design choices.} 
    ``Ours (PETRG-3D)'' is the final model settings. 
    ``PET Enc. (from Scratch)'' trains a separate ViT-3D for the PET branch from scratch. 
    ``w/o Stop Token'' removes the explicit end-of-report token during training. 
    ``w/ Regional Input'' uses 4 decoupled anatomical region pairs for training. 
    ``w/ Regional Input + FT Encoder'' further fine-tunes the visual encoder on top of the ``w/ Regional Input'' setting.
  }
  \label{fig:fig5}
\end{figure}

We conduct further analysis on key design choices, summarized in Figure~\ref{fig:fig5}. ``Ours (PETRG-3D)'' represents our final model settings.

\textbf{CT-Pretrained Features for PET.} Although PET and CT represent distinct modalities, CT-pretrained 3D encoders transfer surprisingly well. Fig.~\ref{fig:fig5} (``PET Enc. (from Scratch)'') shows that our approach—using the CT-pretrained encoder for both modalities—significantly outperforms training a separate PET encoder from scratch. This is likely because anatomical priors, such as tissue texture and lesion morphology, serve as useful inductive biases for PET interpretation.

\textbf{Stop Token.} We find that models without an explicit stop token often produce hallucinated text after the report's factual end. As shown in Fig.~\ref{fig:fig5} (``w/o Stop Token''), adding a dedicated end-of-report token during training significantly improves report faithfulness.

\textbf{Region-wise Decomposition.} We evaluated a regional decomposition strategy: dividing full-body scans and reports into four anatomical regions (e.g., chest, abdomen), training on region-report pairs, and concatenating outputs during inference. This ``w/ Regional Input (RI)'' approach (Fig.~\ref{fig:fig5}) reduces NLG performance. Results indicate that decomposition-induced sample proliferation compromises global contextual coherence critical for radiology report generation.

\textbf{Fine-tuning the Visual Encoder.} We investigate fine-tuning the pretrained CT encoder on our dataset (Fig.~\ref{fig:fig5}, ``w/ RI + FT Encoder''). While this yields modest gains in PET-centric CE (PT-Ab), it degrades both NLG and CT-CE metrics. This suggests that fine-tuning on limited data adapts the model to PET but harms the robust, pre-trained CT representations due to overfitting.

\section{Conclusion}
We introduce PETRG-Lym, a multi-center, single-disease dataset for PETRG, comprising 245,509 paired PET/CT slices and 824 expert-written reports. To facilitate reproducible and fair benchmarking, we also release AutoPET-RG-Lym, a curated PETRG dataset built upon the public AutoPET dataset, with reports carefully drafted and reviewed by domain experts. Based on these datasets, we propose PETRG-3D, an end-to-end 3D PETRG framework featuring a dual-branch architecture for modality-specific encoding and a style-adaptive multimodal fusion module for cross-center stylistic adaptation. Extensive experiments demonstrate that PETRG-3D consistently outperforms existing methods on both NLG and CE metrics, establishing a strong baseline for future research in automated PETRG.

\textbf{Limitations and Future Work:} While this work establishes a complete pipeline—spanning data, modeling, and evaluation—for PETRG, several limitations remain. First, the diagnostic accuracy of generated reports is still below clinical requirements; the model primarily captures stylistic patterns rather than performing detailed pathophysiological reasoning. Second, although our data includes both baseline and post-treatment scans, our framework treats them uniformly. Modeling longitudinal changes (e.g., post-surgical alterations or therapy response) is a crucial next step. In future work, we plan to explore disease-aware reasoning, temporal report synthesis, and clinically validated evaluation protocols to further bridge the gap between research systems and real-world clinical deployment.

{
    \small
    \bibliographystyle{ieeenat_fullname}
    \bibliography{main}
}
\clearpage
\setcounter{page}{1}
\setcounter{section}{0}
\setcounter{figure}{0}
\setcounter{table}{0}

\renewcommand{\thesection}{\Alph{section}}
\maketitlesupplementary

\section{Dataset Details}\subsection{Data Collection}\textbf{PETRG-Lym Acquisition Protocol:} Patients fasted for at least 6 hours prior to the examination. Venous access was established, and blood glucose levels were measured to ensure protocol compliance ($<11.1$ mmol/L). Following the intravenous injection of the $^{18}$F-FDG radiotracer, patients rested in a supine position for approximately 40 minutes. Patients were instructed to void their bladders before scanning to minimize pelvic artifacts. Whole-body (WB) imaging was performed with the patient supine and immobilized, with a typical scan duration of 22 minutes. Attenuation correction (AC) was applied to PET images using the corresponding CT data. Data were acquired using four scanner models: GE Discovery STE, SIEMENS Biograph Vision 450, UNITED IMAGING uMI 510, and uMI Panorama 35S. Diagnostic reports, stored in PDF and DOCX formats, were rigorously reviewed by senior nuclear medicine physicians. This study received ethical approval from the Institutional Review Boards (IRB) of all four participating medical centers.

\subsection{Preprocessing Details}
\label{sec:data details}
\begin{figure*}[t]
  \centering
  \includegraphics[width=0.8\linewidth]{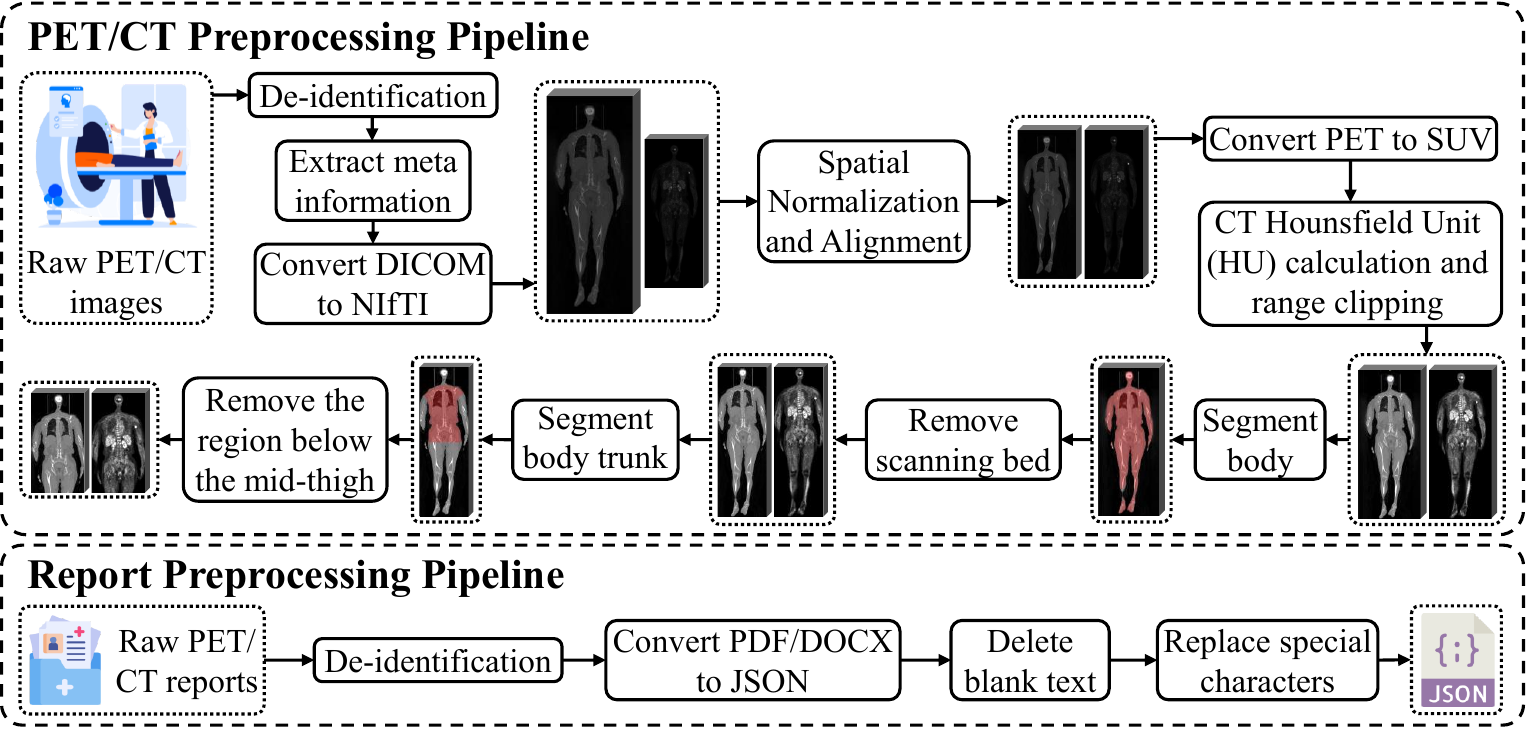}
  \caption{Preprocessing pipeline for PET/CT images and reports.}
  \label{fig:sup fig1}
\end{figure*}

Transforming raw clinical data into model-ready inputs involves a complex pipeline to ensure multi-modal alignment and quantitative consistency. Figure \ref{fig:sup fig1} systematically illustrates our workflow.

\subsubsection{PET/CT Image Preprocessing}
\label{sec:data details}
% --- END FIGURE ---

We implemented a streamlined pipeline to standardize the visual input:
\begin{itemize}
\item \textbf{De-identification:} All Protected Health Information (PHI) was stripped from DICOM headers and file paths.
\item \textbf{Meta-Information Extraction:} Critical metadata—including patient weight, injected dose, acquisition time, and rescale factors—were parsed for quantitative calculations.
\item \textbf{Format Conversion:} Raw DICOM series were converted to NIfTI format for efficient I/O during training.
\item \textbf{Spatial Normalization and Alignment:} Raw PET and CT images often possess differing resolutions and fields of view. To achieve pixel-level alignment, we resampled both modalities to a unified voxel spacing of $1.5 \times 1.5 \times 3$~mm. Both volumes were reoriented to the standard RAS (Right-Anterior-Superior) coordinate system, ensuring strictly aligned multi-modal inputs.
\item \textbf{SUV Calculation:} We converted raw PET intensity to Standardized Uptake Values (SUV) to mitigate inter-subject variability:
\begin{equation}
    \text{SUV}(t) = \frac{c_{img}(t) \cdot \text{BW}}{\text{ID}}
\end{equation}
where $c_{img}(t)$ is the decay-corrected radioactivity concentration (Bq/mL) at scan time $t$, $\text{BW}$ is the body weight (g), and $\text{ID}$ is the injected dose (Bq) corrected to the injection time.

\item \textbf{CT Intensity Normalization:} CT voxel values were converted to Hounsfield Units (HU) using the rescale slope and intercept \cite{dance2014diagnostic}. We clipped the intensities to the range $[-1000, 1000]$ HU to cover the full dynamic range from lung tissue to bone, followed by min-max normalization to $[0, 1]$.

\item \textbf{Background Removal:} To eliminate irrelevant background noise (e.g., scanning bed), we utilized TotalSegmentator~\cite{wasserthal2023totalsegmentator} to generate a precise body mask. The volume was cropped to the minimal bounding box of the body with a 10-slice safety margin.

\item \textbf{Anatomical Focus (ROI Cropping):} Clinical scan ranges vary (e.g., "head-to-toe" vs. "head-to-mid-thigh"). Since lymphoma reporting focuses primarily on the trunk, we standardized the field of view by cropping images to the upper thigh level. Specifically, we retained the body trunk and an extension of 20\% below the pelvic floor ($\leq$ 50 slices) to ensure complete coverage of the inguinal region while discarding irrelevant lower limb data.
\end{itemize}

\subsubsection{Report Text Preprocessing}
The textual data underwent a four-step cleaning process:
\begin{itemize}
\item \textbf{De-identification:} PHI was strictly removed from both raw content and metadata.
\item \textbf{Structure Parsing:} Unstructured reports were parsed into JSON format, extracting key fields: Gender, Clinical History, Findings, and Impression.
\item \textbf{Text Normalization:} Redundant whitespace (e.g., sequences exceeding two spaces) was removed to compact the sequence length.
\item \textbf{Token Standardization:} Special characters were replaced to prevent tokenizer failures (e.g., Roman numerals were converted to Arabic numerals to unify representation while preserving standard staging semantics where possible.), ensuring compatibility with the LLM vocabulary.

\end{itemize}

\section{Implementation Details}
\label{sec:implementation}

\subsection{Implementation of PETRG-3D}

\noindent\textbf{Model Architecture.}
For the PETRG-3D framework, we utilize the ViT3D pre-trained in \cite{wu2023towards} as the volume encoder. 
% --- 修改开始：更专业的Perceiver/Projection描述 ---
We fine-tune the subsequent Perceiver Resampler, which utilizes 128 learnable latent queries to compress the volumetric features into a fixed sequence of 128 visual tokens (with a hidden dimension of 768). These visual tokens are then projected via a linear layer to align with the embedding dimension of the LLM (i.e., 4096).
% --- 修改结束 ---
For the text decoder, we benchmark six state-of-the-art Large Language Models (LLMs) with approximately 8 billion parameters: Llama2-7B\cite{touvron2023llama}, Mistral-7B-v0.3\cite{albert2023mistral}, Qwen2.5-7B\cite{Qwen2024Qwen2}, Gemma2-9B\cite{team2024gemma}, GLM4-9B\cite{glm2024chatglm}, and Qwen3-8B\cite{yang2025qwen3}. 
Given that our target reports are in Chinese, we employ the official pre-trained weights for models with native Chinese support (i.e., Qwen2.5-7B\footnote{\url{https://huggingface.co/Qwen/Qwen2.5-7B-Instruct}}, GLM4-9B\footnote{\url{https://huggingface.co/zai-org/GLM-4-9B-0414}}, and Qwen3-8B\footnote{\url{https://huggingface.co/Qwen/Qwen3-8B}}). Conversely, for Llama2-7B\footnote{\url{https://huggingface.co/FlagAlpha/Llama2-Chinese-7b-Chat}}, Mistral-7B-v0.3\footnote{\url{https://huggingface.co/shenzhi-wang/Mistral-7B-v0.3-Chinese-Chat}}, and Gemma2-9B\footnote{\url{https://huggingface.co/shenzhi-wang/Gemma-2-9B-Chinese-Chat}}, we utilize versions that have been fine-tuned on Chinese corpora to ensure linguistic capability. 

% 一个PET/CT图像报告局部配对的例子
\begin{figure*}[t]
  \centering
  \includegraphics[width=0.98\linewidth]{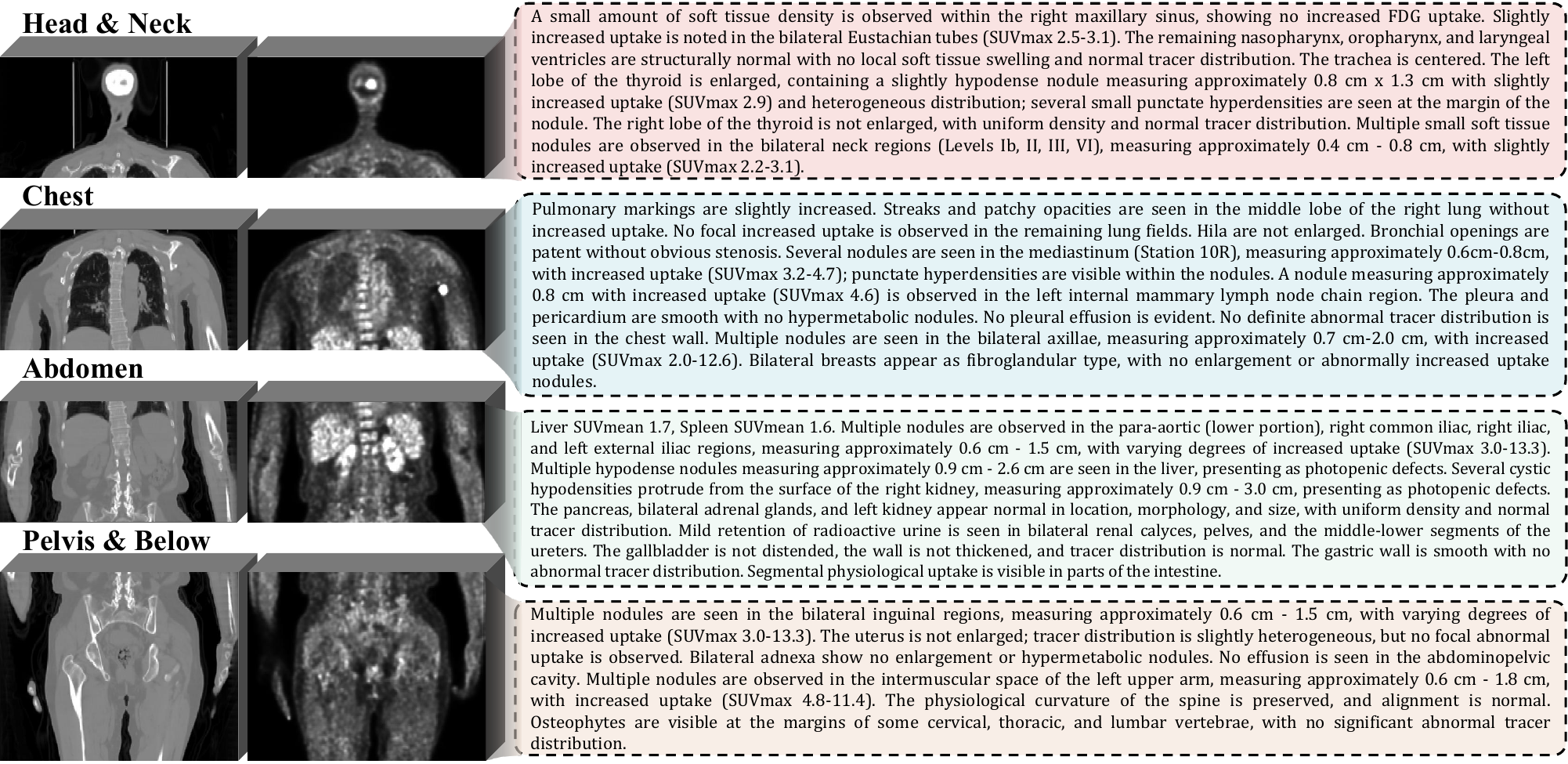}
  \caption{Illustration of region-level image-report pairs in PET/CT.}
  \label{fig:sup fig2}
\end{figure*}
% --- END FIGURE ---

\vspace{0.5em}
\noindent\textbf{Training and Inference.}
% (保持你原有的参数设置不变)
Our framework is implemented in PyTorch 2.8 and trained using the AdamW optimizer. The initial learning rate is initialized at $5\times10^{-5}$ with a linear warmup over the first 100 steps, followed by a constant schedule. 
For parameter-efficient fine-tuning via LoRA, we configure the rank $r=8$, scaling factor $\alpha=32$, and a dropout rate of 0.1. 
The model is trained for 30 epochs on two NVIDIA A800 GPUs with an effective batch size of 16. 
Regarding the LLM configuration, the maximum input sequence length is set to 2048 tokens.  
For evaluation, we employ the model checkpoint from the final training step. During inference, we utilize nucleus sampling with a top-$p$ of 0.9 and a sampling temperature of 0.7. The repetition penalty is set to 1.05, and the maximum generation length is constrained to 1024 tokens.

\vspace{0.5em}
\noindent\textbf{Center-Specific Templates and Explicit Stop Tokens.}
We collected healthy patient report templates from four distinct medical centers, curated by senior nuclear medicine physicians at each institution.
Since drafting PET/CT reports based on standardized healthy templates is a routine workflow in nuclear medicine departments, our Structure-Aware Multi-template Fusion (SAMF) module demonstrates strong generalization capabilities. It can be readily adapted to new centers without the need to reconstruct healthy templates from scratch, as these are typically pre-existing clinical assets.
To explicitly define report termination during training, we append a special token ``[end-of-report]'' to the end of each ground-truth report.

\subsection{Implementation of Comparative Methods}

Due to the scarcity of methodologies dedicated specifically to PETRG, we benchmark our approach against state-of-the-art report generation models from the X-Ray and CT domains.

For methods supporting 3D inputs, namely RadFM~\cite{wu2023towards} and M3D~\cite{bai2024m3d}, we utilize their official implementations and pre-trained weights, adhering to the inference parameters specified in the original papers. As these models are limited to single-modality input, we feed the PET volume during inference for the PETRG task. Furthermore, to address their lack of optimization for Chinese output, we prompt these models to generate reports in English, which are subsequently translated into Chinese using the Qwen3-Max model. 
We manually verified a subset of the translated reports to ensure clinical accuracy and terminology consistency.

Regarding PET2Rep~\cite{zhang2025pet2rep}, we evaluate two slice processing strategies proposed in their work: separate PET and CT slice inputs (PET2Rep-Sep) and fused PET/CT image inputs (PET2Rep-Fus). We perform inference using the official prompts and code. To ensure accurate evaluation, we post-process the raw outputs to filter out artifact symbols (e.g., ``[]'', ``**'') inherent to the template predictions.
For the Vision-Language Model (VLM) backbones, we select the Qwen-VL and InternVL series, which demonstrated superior performance in~\cite{zhang2025pet2rep}. Notably, we employ more recent iterations compared to the original study, utilizing Qwen3-VL-8B and InternVL-3.5-8B. For closed-source models, we employ Qwen-2.5-VL-Max (as Qwen-3-VL-Max was unavailable). 
Finally, for Llava-Med~\cite{nguyen2024multilingual}, we utilize the official code and pre-trained weights. Due to its 2D input constraint, we adopt the fused PET/CT image input strategy from~\cite{zhang2025pet2rep} for inference.

\subsection{Implementation of Region-wise Decomposition}

Clinical PET/CT reports typically follow a strictly ordered anatomical description: head/neck, chest, abdomen, and pelvis. This structural consistency allows for the decomposition of the full report into four distinct regional descriptions. Consequently, we propose a data augmentation strategy that partitions whole-body PET/CT images into these four anatomical sections and pairs them with the corresponding decoupled text (Figure~\ref{fig:sup fig2}). This approach scales the training dataset and simplifies the generative task by focusing on local regions.

\begin{figure}[t]
  \centering
  \includegraphics[width=1.\linewidth]{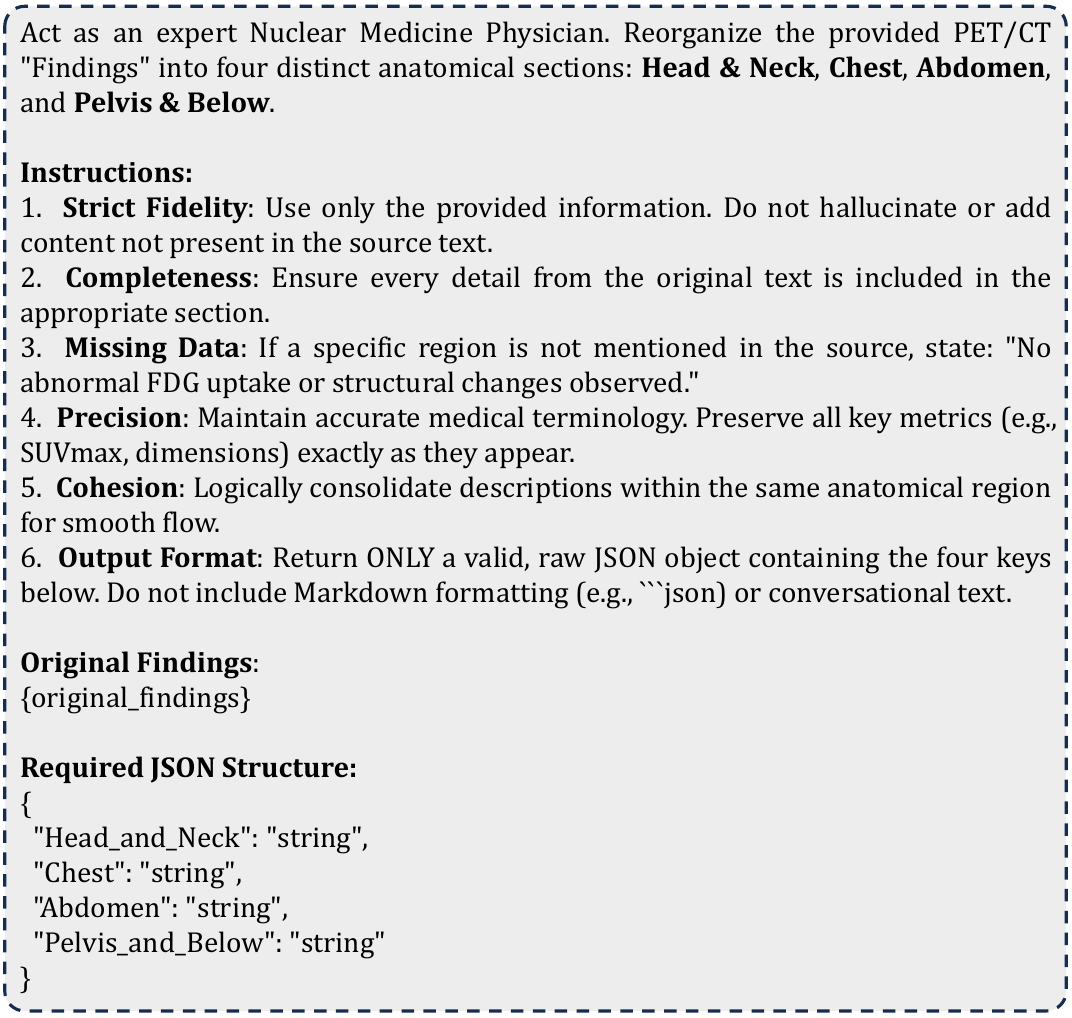}
  \caption{
    Illustration of the prompt design utilized to instruct Qwen3-Max for the anatomical partitioning of whole-body PET/CT reports.
  }
  \label{fig:sup fig3}
\end{figure}

Specifically, we employ a Body Part Regression (BPR) model~\cite{yan2018unsupervised, schuhegger2021mic} to map the coronal slice indices of the CT images to anatomical labels (head, shoulder-neck, chest, abdomen, pelvis, and legs). Based on these mappings, we group ``head'' and ``shoulder-neck'' into the \textit{Head \& Neck} region, while ``chest'', ``abdomen'', and ``pelvis/legs'' form the \textit{Chest}, \textit{Abdomen}, and \textit{Pelvis \& Below} regions, respectively. To mitigate potential information loss caused by BPR boundary errors, we include a buffer of 10 redundant slices at the boundaries of each region.
For textual decomposition, we leverage the Qwen3-Max model to segment the full-body report into the four corresponding sections. Figure~\ref{fig:sup fig3} illustrates the prompt template used for this segmentation, and Figure~\ref{fig:sup fig2} visualizes an example of the region-wise splitting. The resulting dataset contains 2,652 regional image-report pairs.

During training, the model takes regional PET/CT images as input and is supervised by the corresponding regional report. During inference, we generate reports for the four regions of the test images independently and concatenate them to form the final output. To prevent evaluation bias arising from potential content alterations during the LLM-based text segmentation, we construct the reference ground truth for evaluation by concatenating the LLM-segmented regions of the validation set.

\section{Evaluation Metrics}
\label{sec:metrics}

\subsection{NLG Metrics}
To assess the textual similarity between generated and ground-truth reports, we employ three standard Natural Language Generation (NLG) metrics: BLEU~\cite{papineni2002bleu}, METEOR~\cite{banerjee2005meteor}, and ROUGE-L~\cite{lin2004rouge}. Given that Chinese text lacks explicit word delimiters, we utilize the Jieba library\footnote{\url{https://github.com/fxsjy/jieba}} for tokenization prior to calculation.

\subsection{Clinical Efficacy Metric: PETRG-Score}

\subsubsection{Motivation}
Evaluating the clinical utility of generated reports is critical. Standard NLG metrics often fail to capture factual correctness in medical findings. Furthermore, metrics designed for Chest X-Ray or CT report generation are inapplicable to PET/CT, as they ignore metabolic information (PET uptake). Recent attempts, such as PET2Rep~\cite{zhang2025pet2rep} and ViMPET~\cite{nguyen2025toward}, exhibit significant limitations:
\begin{itemize}
    \item \textbf{Incomplete Modality Coverage:} PET2Rep neglects structural abnormalities visible in CT, focusing solely on PET.
    \item \textbf{Coarse Granularity:} ViMPET defines only five broad anatomical regions and uses a binary uptake classification. As shown in Tab.~\ref{tab:sup tab1}, real-world reports contain nuanced uptake descriptions (e.g., physiological vs. mild abnormal uptake) and fine-grained anatomical details (e.g., maxillary sinus) that are essential for \textit{accurate diagnosis and staging}.
\end{itemize}

\begin{table*}[htbp]
\small
  \centering
  \caption{Extraction of anatomical locations, uptake statuses, and density descriptions from the report in Figure~\ref{fig:sup fig2}. This exemplifies the semantic richness of PET/CT reports, underscoring the critical inadequacy of existing oversimplified CE metrics.}
    \begin{tabular}{l|p{10.1em}|p{11.035em}|p{19em}}
    \toprule
    \multicolumn{1}{p{5.465em}|}{\textbf{Region}} & \textbf{Anatomical Location} & \textbf{Uptake Status} & \textbf{Density Status} \\
    \midrule
    \midrule
    \multicolumn{1}{l|}{\multirow{4}[8]{*}{Head\&Neck}} & Right maxillary sinus & Showing no increased FDG uptake & A small amount of soft tissue density is observed \\
\cmidrule{2-4}          & Bilateral Eustachian tubes & Slightly increased uptake is noted & - \\
\cmidrule{2-4}          & Left lobe of the thyroid & Slightly increased uptake and heterogeneous distribution & Enlarged, containing a slightly hypodense nodule measuring approx. 0.8 cm x 1.3 cm; several small punctate hyperdensities are seen at the margin \\
\cmidrule{2-4}          & Bilateral neck regions (Levels Ib, II, III, VI) & Slightly increased uptake & Multiple small soft tissue nodules, measuring approx. 0.4 cm - 0.8 cm \\
    \midrule
    \multicolumn{1}{l|}{\multirow{4}[8]{*}{Chest}} & Middle lobe of right lung & Without increased uptake & Streaks and patchy opacities are seen \\
\cmidrule{2-4}          & Mediastinum (Station 10R) & With increased uptake & Several nodules, measuring approx. 0.6 cm - 0.8 cm; punctate hyperdensities are visible within the nodules \\
\cmidrule{2-4}          & Left internal mammary lymph node chain & With increased uptake & A nodule measuring approx. 0.8 cm \\
\cmidrule{2-4}          & Bilateral axillae & With increased uptake & Multiple nodules, measuring approx. 0.7 cm - 2.0 cm \\
    \midrule
    \multicolumn{1}{l|}{\multirow{5}[10]{*}{Abdomen}} & Liver background & - & - \\
\cmidrule{2-4}          & Para-aortic (lower), Right common iliac, Right iliac, Left external iliac & With varying degrees of increased uptake & Multiple nodules observed, measuring approx. 0.6 cm - 1.5 cm \\
\cmidrule{2-4}          & Liver & Presenting as photopenic defects & Multiple hypodense nodules measuring approx. 0.9 cm - 2.6 cm \\
\cmidrule{2-4}          & Right kidney & Presenting as photopenic defects & Several cystic hypodensities protrude from the surface, measuring approx. 0.9 cm - 3.0 cm \\
\cmidrule{2-4}          & Renal calyces, pelves, ureters & Mild retention of radioactive urine & Mild retention of radioactive urine \\
    \midrule
    \multicolumn{1}{l|}{\multirow{3}[6]{*}{Pelvis\&Below}} & Bilateral inguinal regions & With varying degrees of increased uptake & Multiple nodules, measuring approx. 0.6 cm - 1.5 cm \\
\cmidrule{2-4}          & Intermuscular space of left upper arm & With increased uptake & Multiple nodules observed, measuring approx. 0.6 cm - 1.8 cm \\
\cmidrule{2-4}          & Spine (Cervical, Thoracic, Lumbar) & With no significant abnormal tracer distribution & Osteophytes are visible at the margins of some vertebrae \\
    \bottomrule
    \end{tabular}%
  \label{tab:sup tab1}%
\end{table*}%

\begin{table*}[htbp]
\small
  \centering
  \caption{Comparison of attribute coverage between PETRG-Score and existing metrics. Corroborated by the real-world labels in Tab.~\ref{tab:sup tab1}, PETRG-Score demonstrates significantly more comprehensive anatomical coverage and provides a richer taxonomy for both metabolic uptake and structural density states, surpassing current evaluation standards.}
    \begin{tabular}{l|p{14.6em}|p{10.235em}|p{14.265em}}
    \toprule
    \textbf{Metrics} & \multicolumn{1}{c|}{\textbf{Anatomical Locations}} & \multicolumn{1}{c|}{\textbf{Uptake Status}} & \multicolumn{1}{c}{\textbf{Density Status}} \\
    \midrule
    \midrule
    PET2Rep\cite{zhang2025pet2rep} & Cranium and Brain; Eyeballs; Nasal Cavity and Sinuses; Pharynx and Parapharyngeal Space; Palatine Tonsils and Larynx; Salivary Glands and Thyroid; Cervical Lymph Nodes; Lungs and Thoracic Cavity; Mediastinum and Heart; Esophagus; Liver; Gallbladder; Pancreas; Spleen; Kidneys and Adrenal Glands; Gastrointestinal Tract; Prostate/Uterus and Bladder; Abdominal and Pelvic Cavities; Spine and Bones & Increased Uptake; Decreased Uptake;\newline{}Absent Uptake;  Normal & - \\
    \midrule
    ViMed-PET\cite{nguyen2025toward} & Mediastinum; Lung; Abdomen; Axilla; Cervical Region & Increase; Not Increase & Lymph Node; Pulmonary Nodule; Ground-Glass Opacity; Pulmonary Mass; Pleural\newline{}Thickening; Interstitial Thickening; Consolidation; Effusion; Soft Tissue Nodule; Wall Thickening;\newline{}Calcified Nodule; Hypermetabolic Lesion \\
    \midrule
    PETRG-Score (Ours) & Brain, Skull, and Meninges; Orbit  Nasal Cavity, and Paranasal Sinuses; Pharyngeal Spaces, Tonsils, and Larynx; Thyroid Gland and Major Salivary Glands (Parotid, Submandibular); Cervical Lymph Nodes; Lungs and Pleura; Mediastinum and Hila (including Lymph Nodes); Heart and Pericardium; Axilla and Chest Wall; Breasts; Liver; Gallbladder and Biliary Tract; Spleen; Pancreas; Kidneys; Adrenal Glands; Gastrointestinal Tract (Esophagus, Stomach, Intestines); Retroperitoneal Space (including Lymph Nodes); Peritoneum, Mesentery, and Omentum; Pelvic Organs (Bladder, Uterus/Adnexa or Prostate/Seminal Vesicles); Pelvic and Inguinal Lymph Nodes; Spine; Pelvis and Bones of Extremities; Muscles and Subcutaneous Tissue & Intense Abnormal Uptake; Mild/Suspicious Abnormal Uptake; Physiological/Background Uptake; Uptake Defect / Decreased Uptake; Normal & Lymphadenopathy; Focal Lesion; Lung Parenchymal Abnormality; Wall/Membrane Thickening; Calcification; Bone/Skeletal Lesion; Other Abnormality; Normal \\
    \bottomrule
    \end{tabular}%
  \label{tab:sup tab2}%
\end{table*}%

\subsubsection{Ontology Construction}
To overcome these deficiencies, we established a fine-grained evaluation ontology. We extracted 210 raw anatomical mentions from lymphoma reports, which were reviewed and merged by senior nuclear medicine physicians into a set of 24 clinically significant anatomical regions ($\mathcal{L}$). Similarly, we constructed a comprehensive vocabulary for PET uptake statuses and CT density phenotypes.

\subsubsection{Calculation Pipeline}
\begin{figure*}[t]
  \centering
  \includegraphics[width=1.\linewidth]{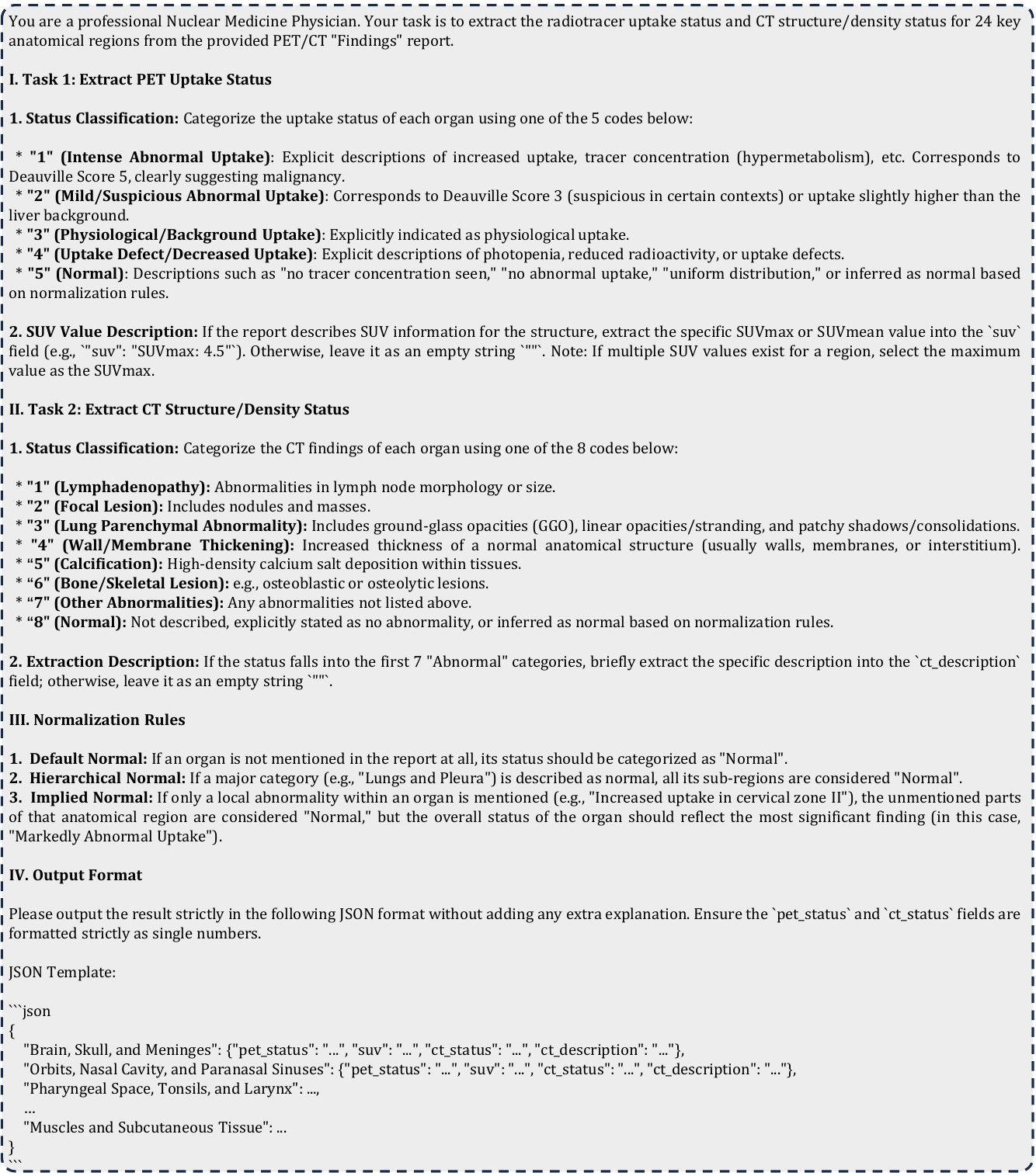}
  \caption{
    Prompt designed for Qwen3-Max to extract attributes from ground-truth and generated reports. The prompt targets 5 uptake states and 8 density states across 24 anatomical locations. We explicitly constrain the output to numerical indices rather than textual labels; this design choice minimizes model hallucinations and ensures robust downstream parsing.
  }
  \label{fig:sup fig4}
\end{figure*}

Based on this expert-defined ontology, we propose the \textbf{PETRG-Score}, a comprehensive metric evaluating both metabolic and structural alignment. The pipeline proceeds as follows:

\noindent \textbf{(1) LLM-based Extraction:} We utilize \textbf{Qwen3-Max} to extract structured labels (anatomical location, uptake status, density status) from both ground-truth and generated reports. To handle linguistic variations and report sparsity, we implement strict \textbf{Normalization Rules} within the prompt (see Fig.~\ref{fig:sup fig4}):
\begin{itemize}
    \item \textit{Default Normal:} Anatomical regions not mentioned in the report are automatically categorized as ``Normal''.
    \item \textit{Hierarchical Inference:} If a major region (e.g., Lungs) is described as normal, all its sub-regions inherit the ``Normal'' status.
    \item \textit{Significance Prioritization:} For regions with local abnormalities, the overall status reflects the most significant finding, while unmentioned sub-parts remain ``Normal''.
\end{itemize}

\noindent \textbf{(2) Expert Verification:} To ensure benchmark reliability, the extracted labels for the ground-truth reports were manually audited and corrected by a researcher specifically trained in medicine imaging.

\noindent \textbf{(3) Metric Formulation:} 
Let $\mathcal{D}$ denote the test set consisting of $N$ reports. For the $j$-th report ($j \in \{1, \dots, N\}$) and the $l$-th anatomical region ($l \in \mathcal{L}$), let $y_{j,l} \in \mathcal{C}$ and $\hat{y}_{j,l} \in \mathcal{C}$ denote the ground-truth and predicted class labels, respectively. The valid category sets are $\mathcal{C}_{CT} = \{1, \dots, 8\}$ and $\mathcal{C}_{PET} = \{1, \dots, 5\}$.

We treat every anatomical region in every report as an individual classification instance. To mitigate the impact of class imbalance, we employ the Macro-F1 score. We first define the True Positives ($TP_k$), False Positives ($FP_k$), and False Negatives ($FN_k$) for a specific category $k \in \mathcal{C}$ by aggregating across all reports and regions:

\begin{equation}
    TP_k = \sum_{j=1}^{N} \sum_{l \in \mathcal{L}} \mathbb{I}(y_{j,l} = k \land \hat{y}_{j,l} = k)
\end{equation}
\begin{equation}
    FP_k = \sum_{j=1}^{N} \sum_{l \in \mathcal{L}} \mathbb{I}(y_{j,l} \neq k \land \hat{y}_{j,l} = k)
\end{equation}
\begin{equation}
    FN_k = \sum_{j=1}^{N} \sum_{l \in \mathcal{L}} \mathbb{I}(y_{j,l} = k \land \hat{y}_{j,l} \neq k)
\end{equation}
where $\mathbb{I}(\cdot)$ is the indicator function.

The F1 score for category $k$ is calculated as: 
\begin{equation}
    F1_k = \frac{2 \cdot TP_k}{2 \cdot TP_k + FP_k + FN_k}
\end{equation}

The final \textbf{PETRG-Score} is the macro-average over a target subset of categories $\mathcal{S} \subseteq \mathcal{C}$:
\begin{equation}
    \text{PETRG-Score}(\mathcal{S}) = \frac{1}{|\mathcal{S}|} \sum_{k \in \mathcal{S}} F1_k
\end{equation}

We report four variants to provide a comprehensive view:
\begin{itemize}
    \item \textbf{PET-All / CT-All:} Calculated over all categories ($\mathcal{S} = \mathcal{C}_{PET}$ or $\mathcal{S} = \mathcal{C}_{CT}$).
    \item \textbf{PET-Ab / CT-Ab:} Calculated only on abnormal categories, excluding the ``Normal'' class (ID: 5 for PET, ID: 8 for CT).
\end{itemize}

\subsubsection{Remarks on Limitations}
\textbf{Extraction Bias:} Verifying labels for all generated reports in ablation studies is prohibitively expensive. We limit manual verification to the ground truth. To ensure fairness, all methods share the same LLM extraction pipeline, ensuring any extraction noise affects all baselines equally.
\textbf{Long-tail Coverage:} While our ontology is significantly richer than prior arts, extremely rare anatomical variants or atypical descriptions may still be underrepresented. Developing dynamic, open-vocabulary evaluation metrics remains a direction for future work.

% --- 更多结果分析 ---
\section{Additional Results}

\subsection{Center-wise Style Adaptation Analysis}
\begin{table}[htbp]
\small
  \centering
  \caption{Multi-center evaluation of the SAMF module’s impact on adapting to heterogeneous reporting styles. ``w/o'' and ``w/ SAMF'' denote the PETRG-3D baseline without and with the module, respectively. Centers 1–4 belong to the PETRG-Lym dataset, while Center 5 corresponds to AutoPET-RG-Lym. Performance gains are highlighted in green parentheses.}
\resizebox{1.\columnwidth}{!}{ 
    \begin{tabular}{l|c|lll}
    \toprule
    \multicolumn{1}{l}{\textbf{Methods}} & \textbf{Center ID} & \textbf{B-4} & \textbf{MTR} & \textbf{R-L} \\
    \midrule
    \midrule
    \multirow{5}[1]{*}{w/o SAMF} & 1     & 26.31 & 44.47 & 38.67 \\
          & 2     & 46.71 & 50.75 & 54.89 \\
          & 3     & 44.86 & 52.53 & 53.47 \\
          & 4     & 35.69 & 42.35 & 46.62 \\
          & 5     & 8.11  & 28.18 & 24.57 \\
    \midrule
    \multirow{5}[1]{*}{w/ SAMF} & 1     & 26.53 \color{ForestGreen}\scriptsize{(+0.22)} & 43.03 \color{magenta}\scriptsize{(-1.44)} & 41.22 \color{ForestGreen}\scriptsize{(+2.55)} \\
          & 2     & 53.38 \color{ForestGreen}\scriptsize{(+6.67)} & 57.19 \color{ForestGreen}\scriptsize{(+6.44)} & 61.81 \color{ForestGreen}\scriptsize{(+6.92)} \\
          & 3     & 55.84 \color{ForestGreen}\scriptsize{(+10.98)} & 62.00 \color{ForestGreen}\scriptsize{(+9.47)} & 63.82 \color{ForestGreen}\scriptsize{(+10.35)} \\
          & 4     & 43.09 \color{ForestGreen}\scriptsize{(+7.40)} & 49.08 \color{ForestGreen}\scriptsize{(+6.73)} & 53.29 \color{ForestGreen}\scriptsize{(+6.67)} \\
          & 5     & 32.47 \color{ForestGreen}\scriptsize{(+24.36)} & 41.50 \color{ForestGreen}\scriptsize{(+13.32)} & 40.97 \color{ForestGreen}\scriptsize{(+16.40)} \\
    \bottomrule
    \end{tabular}%
}
  \label{tab:sup tab3}%
\end{table}%

To quantitatively evaluate the efficacy of the SAMF module in style adaptation, Table~\ref{tab:sup tab3} details the NLG evaluation metrics across four distinct centers within the PETRG-Lym dataset. Integrating the SAMF module yielded consistent improvements across nearly all metrics for the internal centers, confirming its ability to accommodate the diverse reporting conventions inherent to different medical institutions.

Notably, for the external test center (Center 5, AutoPET-RG-Lym), the inclusion of SAMF resulted in substantial performance gains: BLEU-4 surged by 24.36 points, while METEOR and ROUGE-L saw absolute increases of 13.32 and 16.40 points, respectively. These results suggest that even when confronted with unseen reporting styles, PETRG-3D equipped with SAMF effectively performs style transfer by leveraging the synergy between patient-specific PET/CT images and retrieved style templates from the target center.

\subsection{Performance Analysis by Uptake and Density Status}

\begin{table}[htbp]
\small
  \centering
  \caption{Evaluation of our model across five uptake statuses. ``P'' and ``R'' denote Precision and Recall, respectively. ``Support'' indicates the sample size for each status.}
  \resizebox{1.\columnwidth}{!}{ 
    \begin{tabular}{l|ccc|c}
    \toprule
    \multicolumn{1}{c|}{\textbf{Uptake Status}} & \textbf{P} & \textbf{R} & \textbf{F1} & \textbf{Support} \\
    \midrule
    \midrule
    Intense Abnormal Uptake & 35.55  & 19.93  & 25.54  & 537 \\
    Mild/Suspicious Abnormal Uptake & 15.94  & 11.53  & 13.38  & 347 \\
    Physiological/Background Uptake & 28.85  & 23.08  & 25.64  & 65 \\
    Uptake Defect/Decreased Uptake & 13.92  & 13.25  & 13.58  & 83 \\
    Normal & 77.68  & 87.25  & 82.19  & 2832 \\
    \midrule
    Macro Avg & 34.39  & 31.01  & 32.06  & 3864 \\
    Weighted Avg & 64.09  & 68.43  & 65.71  & 3864 \\
    \bottomrule
    \end{tabular}%
}
  \label{tab:sup tab4}%
\end{table}%

\begin{table}[htbp]
\small
  \centering
  \caption{Evaluation of our model across eight density statuses. ``P'' and ``R'' denote Precision and Recall, respectively. ``Support'' indicates the sample size for each status.}
  \resizebox{1.\columnwidth}{!}{ 
    \begin{tabular}{l|ccc|c}
    \toprule
    \multicolumn{1}{c|}{\textbf{Density Status}} & \textbf{P} & \textbf{R} & \textbf{F1} & \textbf{Support} \\
    \midrule
    \midrule
    Lymphadenopathy & 55.56  & 37.72  & 44.93  & 464 \\
    Focal Lesion & 13.46  & 12.07  & 12.73  & 116 \\
    Lung Parenchymal Abnormality & 68.13  & 66.31  & 67.21  & 187 \\
    Wall/Membrane Thickening & 14.29  & 9.52  & 11.43  & 63 \\
    Calcification & 9.30  & 7.55  & 8.33  & 53 \\
    Bone/Skeletal Lesion & 36.99  & 30.00  & 33.13  & 90 \\
    Other Abnormality & 28.98  & 19.85  & 23.56  & 413 \\
    Normal & 72.08  & 82.08  & 76.75  & 2478 \\
    \midrule
    Macro Avg & 37.35  & 33.14  & 34.76  & 3864 \\
    Weighted Avg & 60.91  & 63.82  & 61.84  & 3864 \\
    \bottomrule
    \end{tabular}%
}
  \label{tab:sup tab5}%
\end{table}%

We further stratified model performance by uptake and density statuses to delineate diagnostic capabilities and potential biases. Table~\ref{tab:sup tab4} and Table~\ref{tab:sup tab5} present the Clinical Efficacy (CE) metrics—Precision (P), Recall (R), and F1 score (F1)—alongside the ``Support'' column, which highlights the class distribution.

The results indicate that the ``Normal'' category significantly outperforms others across both uptake and density classifications. This disparity is largely attributable to the extreme class imbalance, where the prevalence of normal samples biases the model towards predicting anatomical regions as abnormality-free. This distribution skew justifies our adoption of the Macro F1 score as a primary evaluation metric to ensure fair assessment.

Interestingly, apart from the normal class, the ``Physiological/Background Uptake'' category (Table~\ref{tab:sup tab4}) achieved the highest performance despite having the fewest samples. This is likely due to the strong semantic coupling between physiological uptake and specific anatomical structures (e.g., the bladder or kidneys), which simplifies the learning task. Similarly, ``Lung Parenchymal Abnormality'' (Table~\ref{tab:sup tab5}) ranks second to the normal category, presumably benefiting from its distinct localization within the lung fields. These findings highlight that while the model learns strong anatomy-pathology associations, mitigating data imbalance remains a critical frontier for future PET/CT report generation research.

\subsection{Computational Resource Analysis}
\begin{table}[htbp]
\small
  \centering
  \caption{Training resource consumption of PETRG-3D with different LLMs on PETRG-Lym dataset. Memory values indicate the peak GPU memory usage (in GB) across two A800 GPUs.}
    \begin{tabular}{l|c|c}
    \toprule
    \textbf{LLMs} & \textbf{Time (Hours)} & \textbf{Memory (GB)} \\
    \midrule
    \midrule
    Llama2-7B & 5.70  & 30.14  \\
    Mistral-7B-v0.3 & 5.56  & 34.49  \\
    Qwen2.5-7B & 5.70  & 38.27  \\
    Gemma2-9B & 8.62  & 45.23  \\
    GLM4-9B & 8.05  & 41.10  \\
    Qwen3-8B & 7.33  & 38.81  \\
    \bottomrule
    \end{tabular}%
  \label{tab:sup tab6}%
\end{table}%

Table~\ref{tab:sup tab6} reports the training time and peak GPU memory usage for PETRG-3D. Experiments were conducted on two NVIDIA A800 (80GB) GPUs using a standardized software stack\footnote{Environment: Python 3.10, CUDA 12.8, PyTorch 2.8.0, Transformers 4.57.0, PEFT 0.17.1, and DeepSpeed 0.17.5.}. As anticipated, both training duration and memory requirements scale positively with the parameter size of the underlying LLM.

\subsection{Qualitative Analysis (Whole-body Reporting)}
% Case 1 GT
\begin{figure*}[t]
  \centering
  \includegraphics[width=0.97\linewidth]{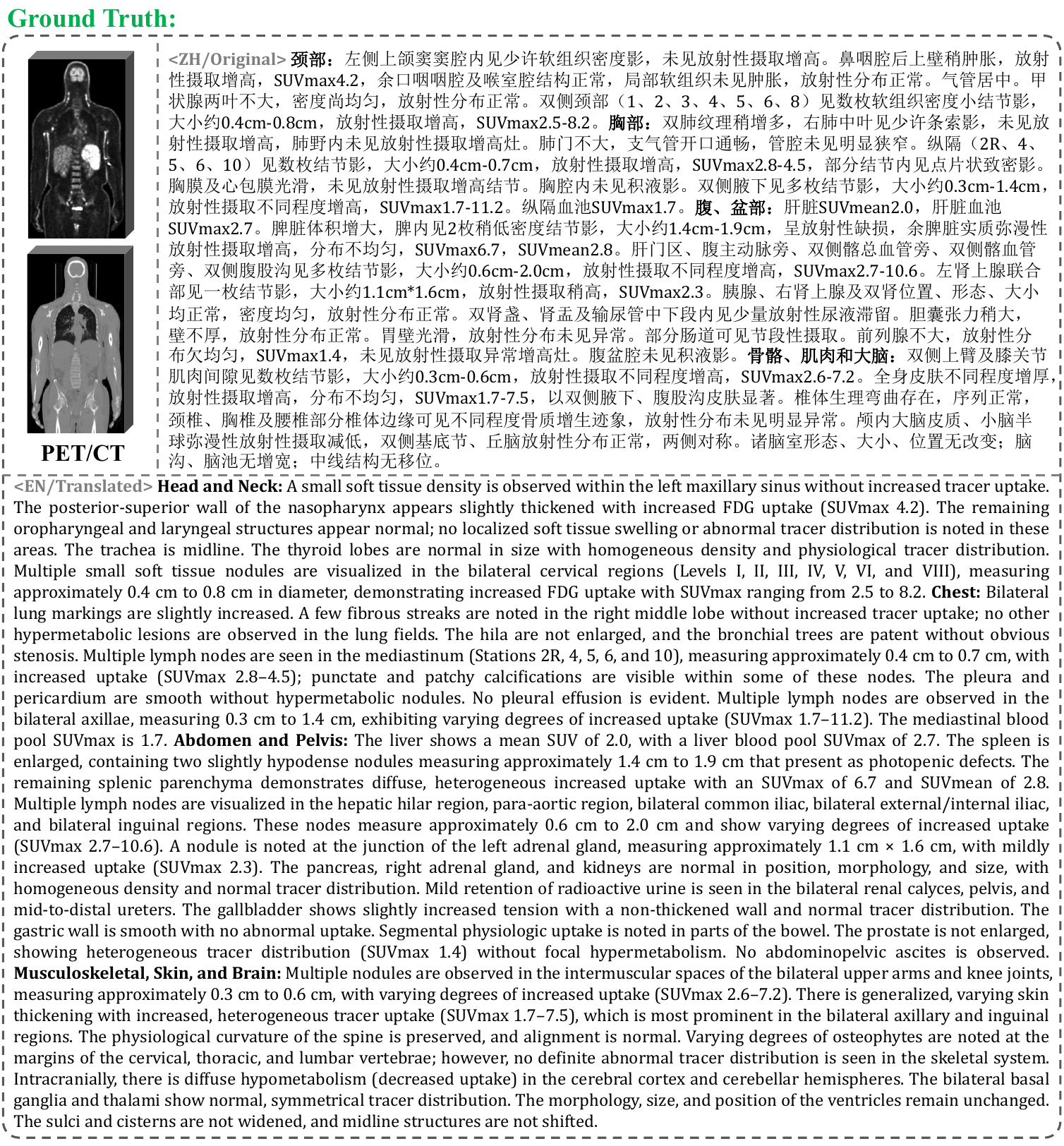}
  \caption{
    Example of a clinical whole-body PET/CT report for lymphoma. The original report is in Chinese (ZH) and has been translated into English (EN) for illustration.
  }
  \label{fig:sup fig5}
\end{figure*}

Figure~\ref{fig:sup fig5} presents a representative whole-body PET/CT lymphoma report. Authored by a nuclear medicine physician, the report is extensive, covering regions from the head and neck to the musculoskeletal system, thereby underscoring the complexity of the PETRG task.

% Case 1 PET2Rep
\begin{figure*}[t]
  \centering
  \includegraphics[width=1.\linewidth]{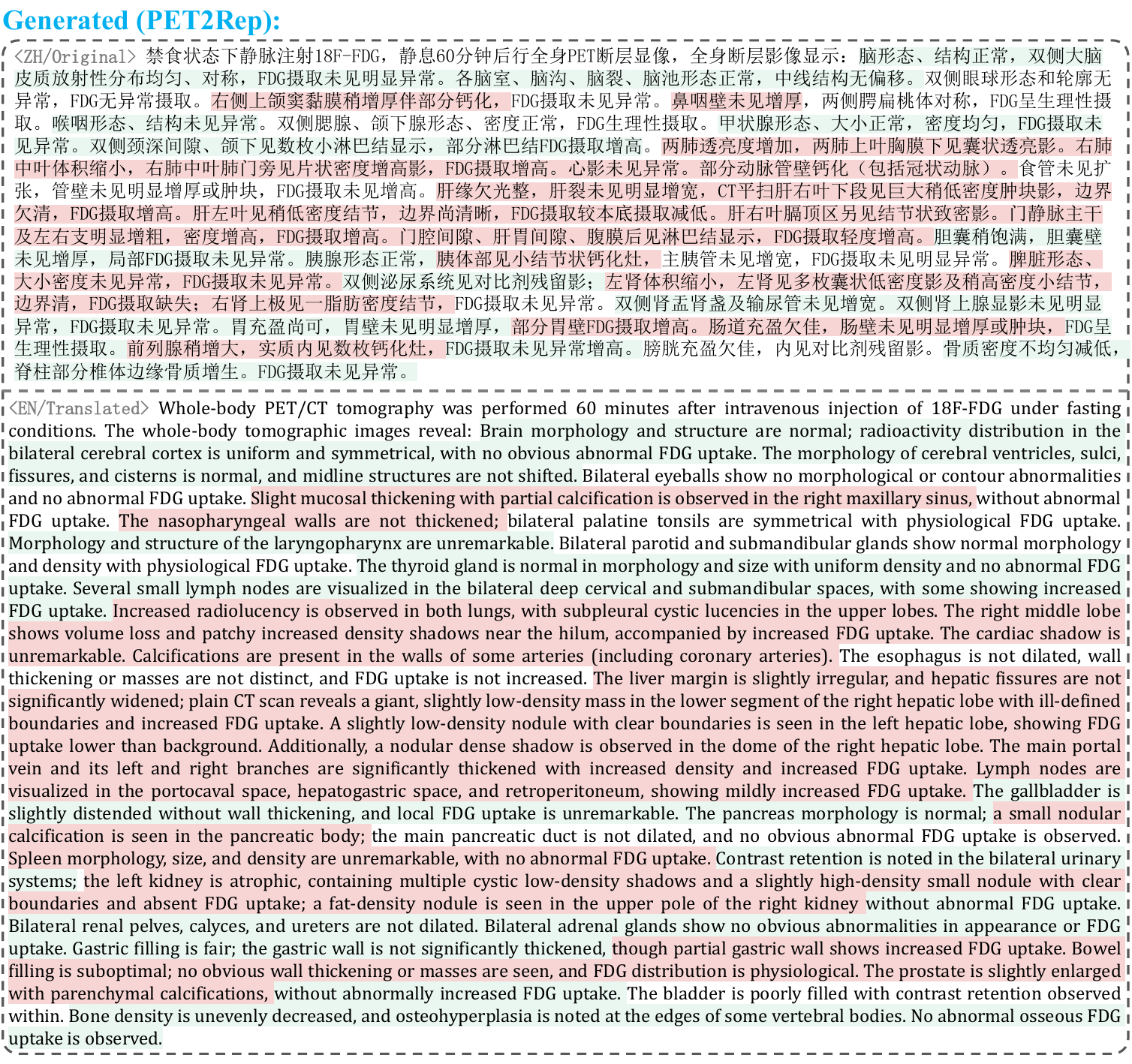}
  \caption{
    Case study on PET2Rep~\cite{zhang2025pet2rep} generation. Green and red backgrounds denote statements consistent and inconsistent with the Ground Truth (Fig.~\ref{fig:sup fig5}), respectively.
  }
  \label{fig:sup fig6}
\end{figure*}

% Case 1 PETRG-3D
\begin{figure*}[t]
  \centering
  \includegraphics[width=1.\linewidth]{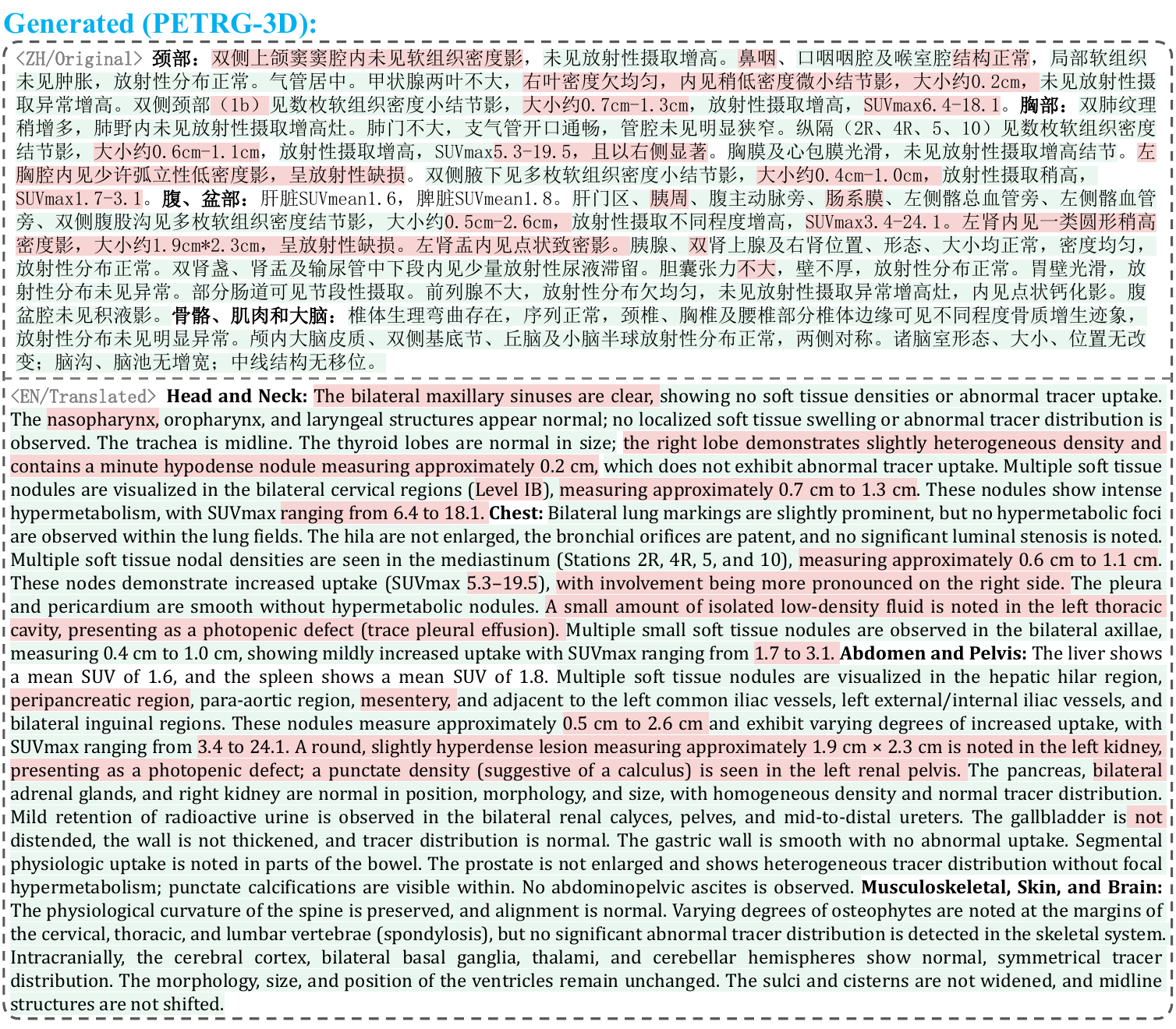}
  \caption{
    Case study on PETRG-3D generation. Green and red backgrounds denote statements consistent and inconsistent with the Ground Truth (Fig.~\ref{fig:sup fig5}), respectively.
  }
  \label{fig:sup fig7}
\end{figure*}

Figure~\ref{fig:sup fig6} illustrates the output from PET2Rep~\cite{zhang2025pet2rep}. As highlighted in red, the baseline model suffers from significant hallucinations, particularly in lesion-rich areas like the lungs and abdomen (e.g., incorrectly predicting volume reduction or pancreatic nodules). This suggests that generalist VLMs or those without specialized fine-tuning lack the diagnostic reliability required for complex medical imaging tasks.

In contrast, the report generated by our PETRG-3D model (Figure~\ref{fig:sup fig7}) closely mirrors the ground truth in both linguistic style and diagnostic content. It accurately identifies multiple nodal involvements (hepatic hilum, para-aortic, iliac, etc.) and extranodal abnormalities. While some limitations remain—specifically regarding quantitative precision (e.g., SUVmax values) and minor false positives in complex regions—the drastic reduction in hallucinations demonstrates the effectiveness of our fine-tuning strategy. Current errors are largely numerical, pointing to future directions for integrating quantitative analysis heads into the architecture.

\subsection{Effectiveness of Explicit Stop Tokens}
\begin{figure*}[t]
  \centering
  \includegraphics[width=1.\linewidth]{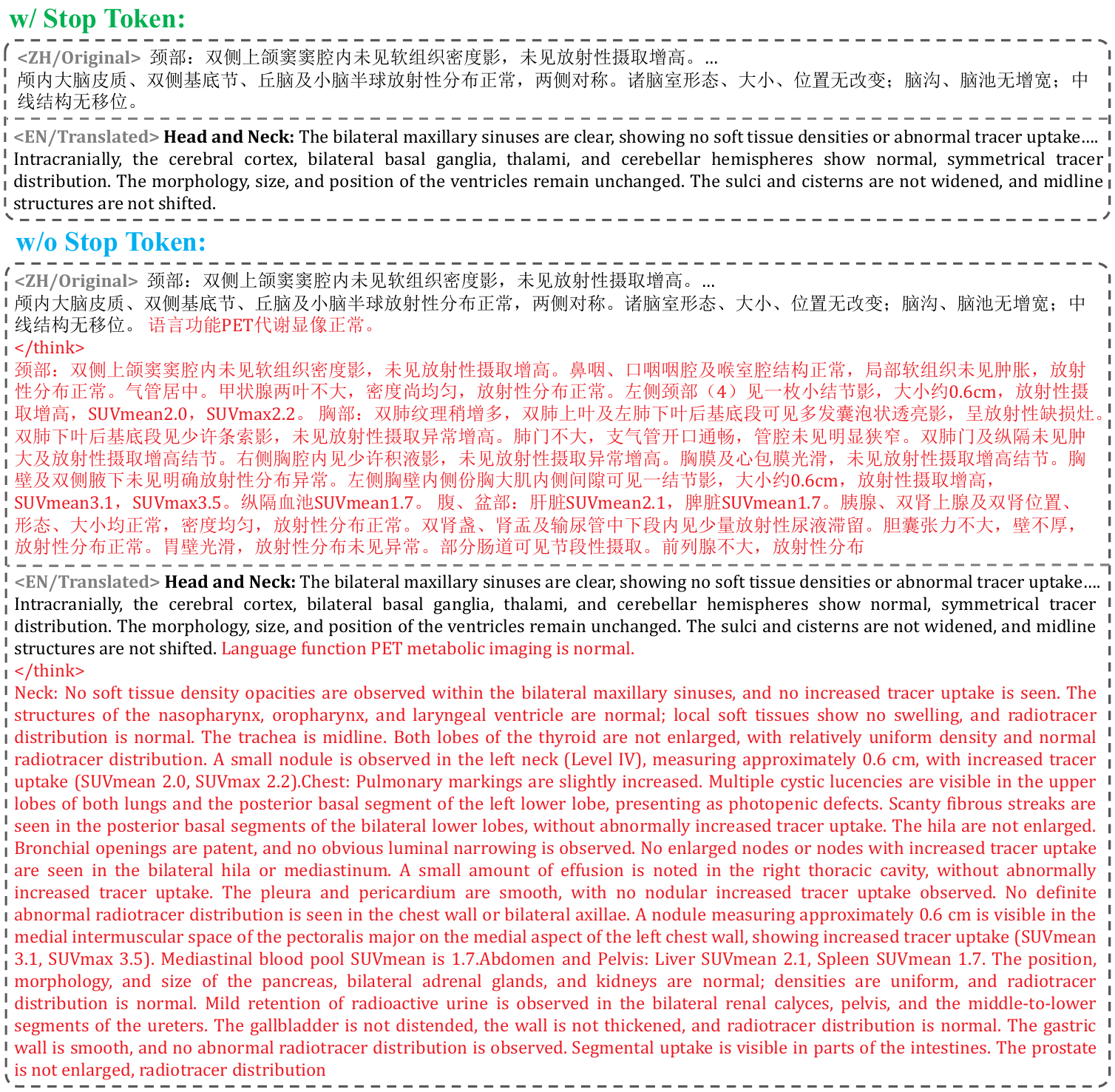}
  \caption{
    Qualitative comparison of generation results with (w/) and without (w/o) the explicit stop token. Note that for the "w/ Stop Token" setting, generation is truncated immediately upon predicting the specific end-of-report token (e.g., ``[end-of-report]''). For brevity, only the beginning and ending sentences are displayed. Red text highlights hallucinations.
  }
  \label{fig:sup fig8}
\end{figure*}

Figure~\ref{fig:sup fig8} compares model outputs before and after the incorporation of explicit stop tokens. Without the explicit stop token, the model tends to generate prolonged hallucinatory content after the actual report concludes, as it attempts to fill the fixed \texttt{max\_new\_token} buffer required for batch generation. By introducing an explicit stop token, the model learns to output ``[end-of-report]'' immediately upon completing the valid report content. This allows for early termination of the generation process, effectively eliminating post-report hallucinations.

\end{document}